\documentclass{article}

\usepackage{times}
\usepackage{pgfplotstable}
\usepackage{comment}

\usepackage{amsmath,amsfonts,bm}

\def\1{\bm{1}}

\def\vzero{{\bm{0}}}

\def\vb{{\bm{b}}}

\def\vd{{\bm{d}}}

\def\vh{{\bm{h}}}

\def\vw{{\bm{w}}}
\def\vx{{\bm{x}}}

\def\vz{{\bm{z}}}
\def\vpi{{\bm{\pi}}}

\def\mA{{\bm{A}}}

\def\mC{{\bm{C}}}

\DeclareMathAlphabet{\mathsfit}{\encodingdefault}{\sfdefault}{m}{sl}
\SetMathAlphabet{\mathsfit}{bold}{\encodingdefault}{\sfdefault}{bx}{n}

\newcommand{\R}{\mathbb{R}}

\def\vzero{{\bm{0}}}

\def\vb{{\bm{b}}}

\def\vd{{\bm{d}}}

\def\vh{{\bm{h}}}

\def\vw{{\bm{w}}}
\def\vx{{\bm{x}}}

\def\vz{{\bm{z}}}

\def\mA{{\bm{A}}}

\def\mC{{\bm{C}}}

\DeclareMathAlphabet{\mathsfit}{\encodingdefault}{\sfdefault}{m}{sl}
\SetMathAlphabet{\mathsfit}{bold}{\encodingdefault}{\sfdefault}{bx}{n}

\usepackage{hyperref}
\usepackage{url}
\usepackage{graphicx}
\usepackage{multirow}
\usepackage{booktabs}
\usepackage{amsmath,amsfonts,amssymb,amsthm}
\usepackage{bm}
\usepackage{eso-pic}
\usepackage{xspace}

\usepackage{tikz}

\usepackage{microtype}
\usepackage{graphicx}
\usepackage{subfigure}
\usepackage{booktabs} 

\usepackage{hyperref}

\usepackage[accepted]{icml2024}

\usepackage{amsmath}
\usepackage{amssymb}
\usepackage{mathtools}
\usepackage{amsthm}

\usepackage[capitalize,noabbrev]{cleveref}

\theoremstyle{plain}

\theoremstyle{definition}

\theoremstyle{remark}

\usepackage[textsize=tiny]{todonotes}

\usetikzlibrary{shapes.geometric, arrows, chains, fit, positioning}

\tikzstyle{var_node} = [circle, minimum width=1cm, minimum height=1cm, text centered, draw=black, fill=red!30]

\tikzstyle{constr_node} = [circle, minimum width=1cm, minimum height=1cm, text centered, draw=black, fill=blue!30]

\tikzstyle{process} = [rectangle, minimum width=1.5cm, minimum height=1cm,rounded corners, text centered, draw=black,text width=1.8cm, fill=yellow!10]

\tikzstyle{data} = [rectangle, minimum width=2cm, minimum height=1cm, rounded corners,text centered, draw=black, text width=2cm,fill=cyan!05]

\tikzstyle{d_data} = [rectangle,  minimum width=3.2cm, minimum height=0.5cm, rounded corners,text centered, draw=black, text width=2cm, fill=cyan!05]

\tikzstyle{mini_data} = [rectangle, minimum width=.5cm, minimum height=1.5cm, rounded corners,text centered, draw=black, text width=.5cm, fill=white!50]

\tikzstyle{huge} = [rectangle,  minimum width=17.5cm, minimum height=5cm, rounded corners,text centered, draw=black, text width=2cm, fill=cyan!05]

\tikzstyle{large} = [rectangle,  minimum width=7cm, minimum height=3.5cm, rounded corners,text centered, draw=black, text width=2cm, fill=cyan!05]

\tikzstyle{medium} = [rectangle,  minimum width=1cm, minimum height=2.5cm, rounded corners,text centered, draw=black, text width=1.5cm, fill=cyan!05]

\tikzstyle{small} = [rectangle,  minimum width=01.4cm, minimum height=0.5cm,text centered, draw=black, text width=1cm, fill=cyan!05]

\tikzstyle{medium_small} = [rectangle,  minimum width=1.4cm, minimum height=.3cm,text centered, draw=black, text width=1cm, fill=white!05]

\tikzstyle{dot} = [circle, minimum width=0.3cm, minimum height=0.3cm, rounded corners,text centered, draw=black,  fill=white!50]

\tikzstyle{blackbox} = [rectangle,  minimum width=3.5cm, minimum height=5.5cm, rounded corners,text centered, draw=black, text width=2cm, fill=orange!30]

\tikzstyle{arrow} = [thick,->,>=stealth, line width=1.5pt]
\tikzstyle{und_arrow} = [thick,--]

\tikzstyle{c_data} = [ellipse, minimum width=1cm, minimum height=6cm,text centered, draw=black, text width=2.5cm,fill=cyan!15]
\tikzstyle{r_flow} = [rectangle,  text width=2.5cm, minimum width=3cm,minimum height=5cm, fill=yellow!15, text centered, draw=black, rounded corners]
\tikzstyle{rr_flow} = [rectangle,  text width=2.5cm, minimum width=3cm,minimum height=5cm, fill=green!15, text centered, draw=black]
\tikzstyle{z_tiny} = [rectangle,  text width=0.25cm, minimum width=0.25cm,minimum height=0.25cm, fill=white!15, text centered, draw=black]
\tikzstyle{r_mlp} = [rectangle,  text width=1.5cm, minimum width=0.75cm,minimum height=0.5cm, fill=white!15, text centered, draw=black]
\tikzstyle{pi_tiny} = [rectangle,  text width=0.6cm, minimum width=0.6cm,minimum height=0.5cm, fill=white!15, text centered, draw=black]

\pgfplotsset{compat=1.18}

\makeatletter
\DeclareRobustCommand\onedot{\futurelet\@let@token\@onedot}
\def\@onedot{\ifx\@let@token.\else.\null\fi\xspace}

\def\eg{\emph{e.g}\onedot} 
\def\ie{\emph{i.e}\onedot} 
\def\cf{\emph{cf}\onedot} 
 
\def\wrt{w.r.t\onedot} 

\makeatother

\icmltitlerunning{Predicting Lagrangian Multipliers for MILPs}

\begin{document}

\twocolumn[
	\icmltitle{Predicting Lagrangian Multipliers for Mixed Integer Linear Programs}




	\icmlsetsymbol{equal}{*}

	\begin{icmlauthorlist}
		\icmlauthor{Francesco Demelas}{lipn}
		\icmlauthor{Joseph Le~Roux}{lipn}
		\icmlauthor{Mathieu Lacroix}{lipn}
		\icmlauthor{Axel Parmentier}{enpc}
	\end{icmlauthorlist}

	\icmlaffiliation{lipn}{Laboratoire d'Informatique de Paris-Nord, Universit\'e Sorbonne Paris Nord --- CNRS, France}
	\icmlaffiliation{enpc}{CERMICS, École des Ponts, France}

	\icmlcorrespondingauthor{Francesco Demelas}{demelas@lipn.fr}
	\icmlcorrespondingauthor{Joseph Le Roux}{leroux@lipn.fr}
	\icmlcorrespondingauthor{Mathieu Lacroix}{lacroix@lipn.fr}

	\icmlkeywords{Machine Learning, ICML, Lagrangian Relaxation, Mixed Integer Linear Programming, Combinatorial Optimization, Bundle Method, Graph Neural Networks}

	\vskip 0.3in
]



\printAffiliationsAndNotice{}  

\begin{abstract}
	Lagrangian Relaxation stands among the most efficient approaches for solving Mixed Integer Linear Programs (MILPs) with difficult constraints.
	Given any duals for these constraints, called Lagrangian Multipliers (LMs), it returns a bound on the optimal value of the MILP, and Lagrangian methods seek the LMs giving the best such bound.
	But these methods generally rely on iterative algorithms resembling gradient descent to maximize the concave piecewise linear dual function:
	the computational burden grows quickly with the number of relaxed constraints.

	We introduce a deep learning approach that bypasses the descent, effectively amortizing \emph{per instance} optimization.
	A probabilistic encoder based on a graph neural network computes, given a MILP instance and its Continuous Relaxation (CR) solution, high-dimensional representations of relaxed constraints, which are turned into LMs by a decoder.
	We train the encoder and the decoder jointly by directly optimizing the bound obtained from the predicted multipliers.
 Our method is applicable to any problem with a compact MILP formulation, and to any Lagrangian Relaxation providing a tighter bound than CR.
	Experiments on two widely known problems, Multi-Commodity
	Network Design and Generalized Assignment, show that our approach closes up to 85~\% of the gap between the continuous relaxation and the best Lagrangian bound, and provides a high-quality warm-start for descent-based Lagrangian methods.
\end{abstract}

\section{Introduction}\label{sec:introduction}

Mixed Integer Linear Programs (MILPs) \citep{wolseyIntegerProgramming2021} have two main strengths that make them ubiquitous in combinatorial optimization \citep{korteCombinatorialOptimizationTheory2012}.
First, they can model many combinatorial optimization problems.
Second, extremely efficient solvers can now handle MILPs with millions of constraints and variables.
They therefore have a wide variety of applications in logistics, telecommunications and beyond.
MILP algorithms are exact: they return an optimal solution, or an optimality gap between the returned solution and an optimal one.

MILPs are sometimes hard to solve due to a collection of difficult constraints.
Typically, a small number of constraints link together otherwise independent subproblems.
For instance, in vehicle routing problems~\citep{golden_vehicle_2008}, there is one independent problem for each vehicle, except for the linking constraints that ensure that exactly one vehicle operates each task of interest.
Lagrangian relaxation approaches are popular in such settings as they allow to unlink the different subproblems.

More formally \citep[Chap.~8]{confortiIntegerProgramming2014}, let $P$ be a MILP of the form:
\begin{subequations}
\begin{align}
	\label{eq:MILP} (P) \qquad& \min_{\vx} \; \vw^{\top}\vx \\
	\label{ctn:ax<=b}  & \mA\vx \ge \vb \\
	\label{ctn:cx<=d}  &  \mC\vx \ge \vd                             \\
	\label{ctn:RxN}    &  \vx \in \mathbb{R}_+^m\times\mathbb{N}^p 
\end{align}
\end{subequations}

While CR amounts to simply removing the integrity constraints (\ie{} \eqref{ctn:RxN} becomes $\vx \in \mathbb{R}_+^{m+p}$),
the relaxed Lagrangian problem is obtained by dualizing difficult constraints~\eqref{ctn:ax<=b} and penalizing their violation with Lagrangian multipliers (LMs) $\vpi \ge \vzero$:

\begin{align*}
	\big(LR (\vpi)\big) \qquad & \min_{\vx} \vw^{\top}\vx+\bm{\pi}^{\top} (\vb-\mA\vx)  \\
	& \mC\vx \geq \vd{} \\
	& \vx{}\in{} \mathbb{R}_+^m\times\mathbb{N}^p
\end{align*}

Standard weak Lagrangian duality ensures that $LR(\vpi)$ is a lower bound on $P$.
The Lagrangian dual problem aims at finding the best such bound:
\begin{equation*}\label{eq:dual}
	(LD)  \qquad \max_{\bm{\pi} \ge \vzero}LR(\bm{\pi}).
\end{equation*}

Geoffrion's theorem~\yrcite{geoffrion_lagrangean_1974} ensures that $LD$ is a lower bound at least as tight as the continuous relaxation. It is strictly better on most applications.
Beyond this bound, Lagrangian approaches are also useful to find good primal solutions.
Indeed, Lagrangian heuristics~\cite{beasley1990lagrangian} exploit the dual solution $\vpi$ and the variable assignment for $\vx$ of  $LR(\vpi)$ to compute good quality solutions of~\eqref{eq:MILP}-\eqref{ctn:RxN}.
Note that both the bound and the heuristic hold even in the case of non-optimal duals $\bm{\pi}$.
We define \emph{good} Lagrangian duals $\bm{\pi}$ as those that lead to a bound $LR(\vpi)$ better than the CR solution, and thus closer to $LD$.

Since $\bm{\pi} \mapsto LR(\vpi)$ is piecewise linear and concave, it is generally optimized using a subgradient algorithm.
Unfortunately, the number of iterations required to obtain good duals quickly increases with the dimension of $\bm{\pi}$, which makes the approach extremely intensive computationally.

In this work\footnote{Code in \textsc{Julia} at \url{https://github.com/FDemelas/Learning_Lagrangian_Multipliers.jl}} we introduce a state-of-the-art encoder-decoder neural network that computes \emph{good} duals $\bm{\pi}$ from the CR solution.
The probabilistic encoder $q_{\bm{\phi}}(\bm{z}|\iota)$, based on a graph neural network (GNN), takes as input a MILP instance $\iota$  as well as the primal and dual CR solutions, and returns an embedding of the instance, where each dualized constraint is mapped to a high-dimensional dense vector.
The deterministic decoder $f_{\bm{\theta}}(\bm{z})$ reconstructs single dimensional duals from constraint vectors.
The learning objective is unsupervised since the Lagrangian dual function $LR(\vpi)$ leads to a natural loss function which does not require gold references.
Experiments on two standard and widely used problems from the Combinatorial Optimization literature, Multi-Commodity Network Design and General Assignment, show that the predicted duals close up to 85\% of the gap between the CR and LD solutions.
Finally, when optimal duals are the target, we show that predicted duals provide an excellent warm-start for state-of-the-art descent-based algorithms for objective~\eqref{eq:dual}.
Our approach is restricted to compact MILPs and Lagrangian Relaxations admitting a tighter bound than CR since primal and dual CR solutions are part of the GNN input.



\section{Learning Framework}\label{sec:learning-framework}

\subsection{Overall Architecture}\label{sec:overall-architecture}

Iterative algorithms for setting LMs to optimality such as the subgradient method (SM)~\citep[Chap 5.3]{Polyak87} or the Bundle method (BM)~\citep{hiriart1996convex2,le2007bundle}
start by initializing LMs.
They can be set to zero but a solution considered as better in practice by the Combinatorial Optimization community is to take advantage of the bound given by CR and its dual solution, often  computationally cheap for compact MILPs.
Specifically, optimal values of the CR dual variables identified with the constraints dualized in the Lagrangian relaxation can be understood as LMs.
In many problems of interest these LMs are not optimal and can be improved by SM or BM.\@
We leverage this observation by trying to predict a deviation from the LMs corresponding to the CR dual solution.

\begin{figure*}
	\centering
	 	\begin{tikzpicture}[node distance=3cm, scale=0.5,transform shape]
  
 		\node (data) [c_data] {\Large MILP $\iota$\\ + \\ CR \\ Solution \\ $\mathbf{x},\pmb{\lambda}$\\+\\  Dualized Constraints};

        \node(data_labels)[below=0.2cm of data.south]{\Large Input};

    \node (encoder) [rectangle,minimum height=6cm,minimum width=10.75cm,right=0.25cm of data, draw=black, fill=green!15] {};
\node(encoder_labels)[below=0.25cm of encoder.south]{\Large Encoder};

 		\node (FE) [r_flow, right=0.5 of data] {\Large Features\\ Extraction};
 		\node (GNN) [r_flow, right=0.5 of FE, minimum width=3.2cm] {};
   
 		\node (gnn1) [pi_tiny, right=0.2 of GNN.west, minimum height=1.5cm]{};
 		\node (gnn_dots) [right=0.1 of gnn1, minimum height=1.5cm]{\Large $\cdots$};
 		\node (gnn1) [pi_tiny, right=0.1 of gnn_dots, minimum height=1.5cm]{};

 		\node (GNN_label) [below=0.5cm of GNN.north] {\Large $\text{GNN}_{\pmb{\phi}}$};

 		\node (Sampler) [r_flow, right=0.5 of GNN] {\Large Sampler};

        \node (zs) [c_data,  right=0.75 of Sampler, yshift=0.cm] {};
        \node (zs_1_1) [z_tiny, below=0.75cm of zs.north] {};
        \node (zs_1_2) [z_tiny, below=0cm of zs_1_1] {};
        \node (zs_1_3) [z_tiny, below=0cm of zs_1_2] {};
        \node (zs_1_4) [z_tiny, below=0cm of zs_1_3] {};
        \node (zs_1_label) [right=0.2 of zs_1_3] {\Large $\mathbf{z_1}$};
        
        \node (zs_2_1) [z_tiny, below=0.2cm of zs_1_4] {};
        \node (zs_2_2) [z_tiny, below=0cm of zs_2_1] {};
        \node (zs_2_3) [z_tiny, below=0cm of zs_2_2] {};
        \node (zs_2_4) [z_tiny, below=0cm of zs_2_3] {};
        \node (zs_2_label) [right=0.2 of zs_2_3] {\Large $\mathbf{z_2}$};

        \node (zs_dots) [below=0.3cm of zs_2_3]{\Large $\mathbf{\vdots}$};

        \node (zs_3_1) [z_tiny, below=0.3cm of zs_dots] {};
        \node (zs_3_2) [z_tiny, below=0cm of zs_3_1] {};
        \node (zs_3_3) [z_tiny, below=0cm of zs_3_2] {};
        \node (zs_3_4) [z_tiny, below=0cm of zs_3_3] {};
        \node (zs_3_label) [right=0.2 of zs_3_3] {\Large $\mathbf{z_{|\lambda|}}$};

        \node(zs_labels)[below=1.3cm of zs_3_4.south]{\Large Latent};

 		\node (decoder) [rr_flow, right=0.5 of zs] {};
        \node(decoder_labels)[below=0.75cm of decoder.south]{\Large Decoder};

        \node (MLP_1) [r_mlp, below=1cm of decoder.north] {\Large ${MLP_{\pmb{\theta}}}$};
        \node (MLP_2) [r_mlp, below=0cm of MLP_1] {\Large ${MLP_{\pmb{\theta}}}$};
        \node (MLP_dots) [ below=0.2cm of MLP_2] {\Large $\mathbf{\vdots}$};
        \node (MLP_3) [r_mlp, below=0.2cm of MLP_dots] {\Large${MLP_{\pmb{\theta}}}$};
   
 		\node (pis) [c_data,    right=0.5 of decoder] {};

        \node (pi_1) [pi_tiny, below=1.6cm of pis.north] {\Large$\pi_1$};
        \node (pi_2) [pi_tiny, below=0cm of pi_1] {\Large$\pi_2$};
        \node (pi_dots) [ below=0.2cm of pi_2] {\Large $\mathbf{\vdots}$};
        \node (pi_3) [pi_tiny, below=0.3cm of pi_dots] {\Large$\pi_{|\lambda|}$};

        \node(pi_labels)[below=1.85cm of pi_3.south]{\Large Multipliers};
   
 		\node (eb) [r_flow, right=0.5 of pis, fill=red!15] {\Large Energy \\ Bound \\ \vspace{1cm} \Large $LR(\pmb{\pi})$};

      \draw [arrow] (zs_1_label) -> (MLP_1);
	   \draw [arrow] (zs_2_label) -> (MLP_2);
	   \draw [arrow] (zs_3_label) -> (MLP_3);
	
      \draw [arrow]  (MLP_1) -> (pi_1);
	   \draw [arrow] (MLP_2)  -> (pi_2);
	   \draw [arrow] (MLP_3) -> (pi_3);

\draw [arrow] (data) -> (FE); 
\draw [arrow] (FE) -> (GNN);
\draw [arrow] (GNN) -> (Sampler);

\draw [arrow] (Sampler) -> (zs);

\draw [arrow] (pis) -> (eb);
 
 \end{tikzpicture}
	\caption{Overall Architecture. From the bipartite graph representation of a MILP and its CR solution, the model computes a Lagrangian dual solution.
		First the MILP is encoded by a GNN, from which we parameterize a sampler for constraint representations.
		These representations are then passed through a decoder to compute Lagrangian Multipliers.
	}\label{fig:model}
\end{figure*}
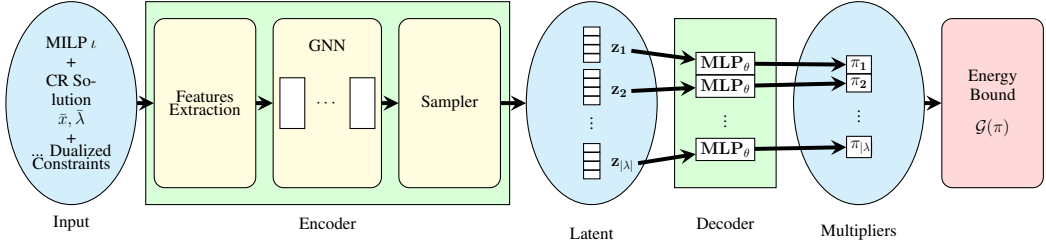

The architecture is depicted in Figure~\ref{fig:model}.
We start from an input instance \(\iota\) of MILP $P$ with a set of constraints for which the Lagrangian relaxed problem is easy to compute, then solve \(CR\) and obtain the corresponding primal and dual solutions.
The input enriched with \(CR\) solutions is then passed through a probabilistic encoder, composed of three parts: \emph{(i)} the input is encoded as a bipartite  graph in a way similar to~\citep{gasse_exact_2019}, also known as a \emph{factor graph} in probabilistic modelling, and initial graph node feature extraction is performed, \emph{(ii)} this graph is fed to a GNN in charge of refining the node features by taking into account the structure of the MILP, \emph{(iii)} the last layer of the GNN is used to parameterize a distribution from which vectors \(\vz_{c}\) can be sampled  for each dualized constraint \(c\).

The decoder then translates $\vz_{c}$ to a positive LM $\pi_{c}=\lambda_{c}+ \delta_{c}$ by predicting a deviation $\delta_{c}$ from the CR dual solution variable $\lambda_{c}$.
Finally, the predicted LMs can be used in several ways, in particular to compute a Lagrangian bound or to warm-start an iterative solver.

\subsection{Objective}\label{sec:objective}

We train the network's parameters in an end-to-end fashion by maximizing the average Lagrangian bound $LR(\vpi)$ obtained from the predicted LMs $\bm{\pi}$ over a training set.
This can be cast as an empirical risk optimization, or an Energy-Based Model~\citep{lecun2006tutorial} with latent variables, where the Lagrangian bound is the (negative) energy corresponding to the coupling of the instance with the subproblem solutions, and the LMs --- or more precisely their high-dimensional representations --- the latent variables.
For our problem, a natural measure of the quality of the prediction is provided by the value $LR$ that we want to maximize to tighten the duality gap.
Given an instance \(\iota\) we want to learn to predict the latent representations $\bm{z}$ of the LMs for which the Lagrangian bound is the highest:

\[
	\max_{\bm{\phi},\bm{\theta}}\mathbb{E}_{\bm{z}\sim q_{\bm{\phi}}(\cdot|\iota)}\left[LR ([\bm{\lambda} + f_{\bm{\theta}}(\bm{z})]_+; \iota)\right]
\]
\noindent where \(q_{\bm{\phi}}\) is the probabilistic encoder, mapping each dualized constraint $c$ in \(\iota\) to a latent vector $\bm{z}_{c}$ computed by independent Gaussian distributions, $f_{\bm{\theta}}$ is the decoder mapping each\footnote{With a slight abuse of notation, we use function  $f:\mathbb{R}^{m}\to \mathbb{R}^{n}$ on \emph{batches} of size $p$ to become $\mathbb{R}^{m\times p}\to \mathbb{R}^{n\times p}$.} $\bm{z}_{c}$ to its corresponding LM deviation \(\delta_{c}\) from the CR dual value $\lambda_{c}$, and \([\cdot]_+\) is the component-wise softplus function. We can observe that this objective has the following properties amenable to gradient-based learning:
\begin{enumerate}
	\item \(LR(\vpi)\) is bounded from above:  optimal LMs $\bm{\pi}^*$ maximize $LR(\vpi)$ over all possible LMs, that is $LR(\bm{\pi}^*)\geq LR(\bm{\pi})$ for any \(\bm{\pi} = \bm{\lambda} + f_{\bm{\theta}}(\bm{z})\).
	      Moreover, \(LR(\vpi)\) is a concave piece-wise linear function, in other words all optimal solutions will give the same bound.
	\item It is straightforward to compute a subgradient \wrt{} to   parameters \(\bm{\theta}\):  \(\nabla_{\bm{\theta}} LR([\bm{\lambda} + f_{\bm{\theta}}(\bm{z})]_+; \iota)\) is equal to:
	      \[
		       \left(\frac{\partial [\bm{\lambda} + f_{\bm{\theta}}(\bm{z})]_+}{\partial \bm{\theta}}\right)^{\top} \nabla_{\bm{\pi}} LR(\bm{\pi}; \iota)
	      \]
	      The Jacobian on the left is computed via backpropagation, while  $LR(\vpi; \iota)$ is simple enough for a subgradient to be given analytically.
	      Provided that $\bar{\bm{x}}$ is an optimal solution of the relaxed Lagrangian problem of $\iota$ associated with  $\bm{\pi}$, we derive:
	      \[
		      \nabla_{\bm{\pi}} LR(\bm{\pi}; \iota) = \bm{b}-\bm{A}\bar{\bm{x}}
	      \]
				This means that in order to compute a subgradient for $\bm{\theta}$, we first need to solve each subproblem.
				Since subproblems are independent, this can be done in parallel.

	\item For parameters \(\bm{\phi}\), we again leverage function composition and the fact that \(q_{\bm{\phi}}\) is a Gaussian distribution, so we can approximate the expectation by sampling and use the reparameterization trick~\citep{Kingma2014, NIPS2015_de03beff} to perform standard backpropagation.
	      We implement $q_{\bm{\phi}}$ as a neural network, described in details in the following section, returning a mean vector and a variance vector for each dualized constraint \(c\), from which a sampler returns a representation vector \(\bm{z}_{c}\).
	      For numerical stability, the variance is clipped to a safe interval~\citep{pmlr-v139-rybkin21a}.
\end{enumerate}

\subsection{Encoding and Decoding Instances}\label{sec:enc-dec}

\paragraph{Encoder} One of the challenges in Machine Learning applications to Combinatorial Optimization is that instances have different input sizes, and so the encoder must be able to cope with these variations to produce high-quality features.
Of course this is also the case in many other applications, for instance NLP where texts may differ in size, but there is no general consensus as to what a good feature extractor for MILP instances looks like, contrarily to other domains where variants of RNNs or Transformers have become the de facto standard encoders.

We depart from previous approaches to Lagrangian prediction~\cite{sugishita_warmstarting_2021} restricted to instances of the same size, and follow more generic approaches to MILP encoding such as~\citep{gasse_exact_2019,smipunn2020,Khalil2017LearningTR} where each instance is converted into a bipartite graph and further encoded by GNNs to compute meaningful feature vectors associated with dualized constraints.
Each MILP is converted to a bipartite graph composed of one node for each variable and one node for each constraint.
There is an edge between a variable node $n_{v}$ and a constraint  node $n_{c}$ if and only if $v$ appears in $c$.
Each node (variable or constraint) is represented by an initial feature vector \(\bm{e}_{n}\).
We use features similar to ones given in~\citep{gasse_exact_2019}.\footnote{See Appendix~\ref{sec:initial-features} for more details.}
Following \citet{smipunn2020}, variables and constraints are encoded as the concatenation of variable features followed by constraint features, of which only one is non-zero, depending on the type of nodes.


To design our stack of GNNs, we take inspiration from structured prediction models for images and texts, where Transformers~\citep{vaswani2017attention} are ubiquitous.
However, since our input has a bipartite graph structure, we replace the multihead self-attention layers with simple linear graph convolutions\footnote{Alternatively, this can be seen as a masked attention, where the mask is derived from the input graph adjacency matrix.}~\citep{kipf2016semi}.
Closer to our work, we follow \citet{smipunn2020} which showed that residual connections~\citep{he2016residual}, dropout~\citep{srivastava2014dropout} and layer normalization~\citep{ba2016layer} are important for the successful implementation of feature extractors for MILP bipartite graphs.

\begin{figure*}
	\centering
		\begin{tikzpicture}[scale=0.7, transform shape] \centering \scriptsize
		\node (hi) [medium,ellipse] { };
		\node (label_hi) [anchor=north, text width=1cm,above = 0.2cm of hi] {\textbf{node}\\ \textbf{features}};

		\node (block) [huge, right =0.5cm of hi,fill=orange!20,yshift=0.25cm] {  };

		\node (label_block) [anchor=north, above = -0.7cm of block] {\large Graph Neural Network Block};

		\node (gcn) [large, right=0.8cm of hi,fill=cyan!40] {  };
		\node (label_gcn) [anchor=north, above = -0.6cm of gcn] {\normalsize Graph Message Passing};

		\node (mlp) [large, right=3cm of gcn, fill=green!30] {};

		\node (label_mlp) [anchor=north,above = -0.5cm of mlp] {\normalsize Non-Linear Transformation };

		\node (s1) [small, above  =-0.9cm of hi,fill=orange!10] {$\vh_1^i$};
		\node (sdots) [medium_small, below=0cm of s1] {\Large $\mathbf{\vdots}$};
		\node (sn) [small, below=0cm of sdots,fill=blue!10] {$\vh_n^i$};

		\node (ln1) [medium, right= 1cm of hi, fill=white!30] { Layer Normalization};
		\node (ln2) [medium, right= 3.25cm of gcn, fill=white!30] { Layer Normalization};

		\node (gcnl) [medium, right= 0.5cm of ln1, fill=red!30] {};
		\node (gcnl_label) [above = -1cm of gcnl,text width=1.5cm] { Graph\\ Convolution Layer};

		\node (d1) [dot, right=0.8cm of ln1,xshift=0.1cm]{};
		\node (d2) [dot, right =0.3cm of d1,yshift=0.1cm]{};
		\node (d3) [dot, below =0.5cm of d1,xshift=0.1cm,xshift=-0.1cm]{};
		\node (d4) [dot, right =0.5cm of d3,yshift=-0.3cm]{};
		\draw (d1) --(d2);
		\draw (d2) --(d3);
		\draw (d3) --(d4);
		\draw (d1) --(d4);

		\node (mlpl) [medium, right= 0.5cm of ln2, fill=yellow!30] {};

		\node (mlpl_label) [above = -0.4cm of mlpl,text width=1.5cm] {Parallel MLP};

		\node (mlp1) [small, above =-1cm of mlpl,fill=green!10] {$MLP_{\bm{\phi}}$};
		\node (mlpdots) [medium_small, below=0cm of mlp1,fill=green!10] {\Large $\mathbf{\vdots}$};
		\node (mlpn) [small, below=0cm of mlpdots,fill=green!10] {$MLP_{\bm{\phi}}$};

		\node (do1) [medium, right= 0.5cm of gcnl] { Dropout\\ Layer};
		\node (do2) [medium, right= 0.5cm of mlpl] { Dropout\\ Layer};

		\node (hfr) [medium, right= 0.5cm of gcn,ellipse] {};
		\node (s1_hfr) [small, above  =-0.9cm of hfr,fill=orange!10] {${\vh_1^{i}}^{'}$};
		\node (sdots_hfr) [medium_small, below=0cm of s1_hfr] {\Large $\mathbf{\vdots}$};
		\node (sn_hfr) [small, below=0cm of sdots_hfr,fill=blue!10] {${\vh_n^i}^{'}$};

		\node (hip1) [medium, right=1.5cm of do2.east,ellipse] {};
		\node (label_hip1) [anchor=north, text width=1cm,above = 0.3cm of hip1] {\textbf{node}\\ \textbf{features}};

		\node (s1p1) [small, above =-0.9cm of hip1,fill=orange!10] {$\vh_1^{i+1}$};
		\node (sdotsp1) [medium_small, below=0cm of s1p1] {\Large $\mathbf{\vdots}$};
		\node (snp1) [small, below=0cm of sdotsp1,fill=blue!10] {$\vh_n^{i+1}$};

		\draw [arrow] (hi) -- (ln1);
		\draw [arrow] (ln1) -- (gcnl);
		\draw [arrow] (gcnl) -- (do1);
	
         \node (c1) [circle,right=-0.6cm of gcn,draw=black,fill=white] {$+$};  
    
        \draw [thick, line width=1.5pt] (do1) -- (c1);
        \draw [arrow] (c1) -- (hfr);
        
		\draw [arrow] (hfr) -- (ln2);
		\draw [arrow] (ln2) -- (mlpl);
		\draw [arrow] (mlpl) -- (do2);
		\draw [arrow] (do2) -- (hip1);

		\node (img1) [below=1.9cm of gcn.east,xshift=-0.425cm] {};
		\draw [thick, line width=1.5pt] (hi.south) |- (img1.east);
		\draw [thick, line width=1.5pt] (img1.east) -- (c1);

		\node (img2) [below=1.9cm of mlp.east,xshift=-0.425cm] {};
		\draw [thick, line width=1.5pt] (hfr) |- (img2.east);
  
         \node (c2) [circle,right=-0.6 of mlp,draw=black,fill=white] {$+$};  
    
        \draw [thick, line width=1.5pt] (do2) -- (c2);
        \draw [arrow] (c2) -- (hip1);
  
		\draw [thick, line width=1.5pt] (img2.east) -- (c2);

	\end{tikzpicture}
	\caption{The Graph Neural Network block.
		The first part is graph message-passing: we apply layer normalization to  node features, then convolution over the instance's bipartite graph representation and finally dropout.
		The second phase consists of normalization, a Multi-Layer perceptron in parallel over all the nodes of the bipartite graph,
		then dropout. Both sublayers use residual connection between input and output.
		We apply this block several times to improve feature representations.
	}\label{fig:transformer_block_model}
\end{figure*}
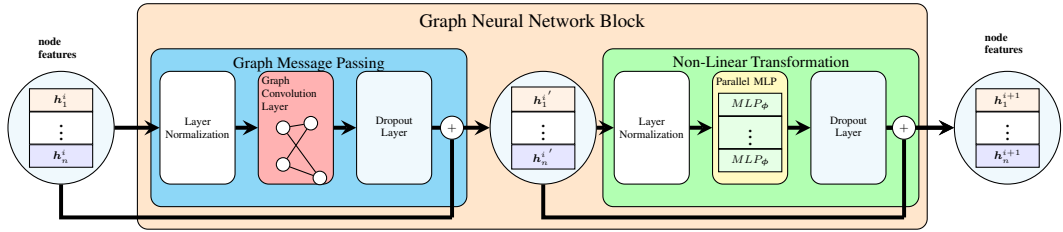

Before the actual GNNs, initial feature vectors \({\{\bm{e}_{n}\}}_{n}\) are passed through a MLP $F$ to find feature combinations and extend node representations to high-dimensional spaces: \(\vh_{n}^{0} = F(\bm{e}_{n}), \forall n\).
Then interactions between nodes are taken into account by passing vectors through blocks, represented in Figure~\ref{fig:transformer_block_model}, consisting of two sublayers. 
\begin{itemize}
	\item The first sublayer connects its input via a residual connection to a layer normalization \(LN\) followed by a linear graph convolution \(CONV\) of length 1, followed by a dropout regularization \(DO\):
	      \[
		      \vh'_{n} = \vh_{n} + DO(CONV(LN(\vh_{n})))
	      \]
	      The graph convolution passes messages between nodes. In our context, it passes information from variables to constraints, and conversely.

	\item The second sublayer takes as input the result of first one, and connects it with a residual connection to a sequence made of a layer normalization \(LN\), a MLP transformation and a dropout regularization \(DO\):
	      \[
		      \vh_{n} = \vh'_{n} + DO(MLP(LN(\vh'_{n})))
	      \]
	      The MLP is in charge of finding non-linear interactions in the information collected in the previous sublayer.

\end{itemize}

This block structure, depicted in Figure~\ref{fig:transformer_block_model}, is repeated several times, typically 5 times in our experiments, in order to extend the domain of locality.
The learnable parameters of a block are the parameters of the convolution in the first sublayer and the parameters of the MLP in the second one.
Remark that we start each sublayer with normalization, as it has become the standard approach in Transformer recently~\citep{chen-etal-2018-best}. We note in passing that this has also been experimented with by~\citet{gasse_exact_2019} in the context of MILP, although only once before the GNN input, whereas we normalize twice per block, at each block.

Finally, the GNN returns the vectors associated with dualized constraints \(\{\vh_{c}\}_{c}\).
Each vector \(\vh_{c}\) is interpreted as the concatenation of two vectors \([\bm{z}_{\mu}; \bm{z}_{\sigma}]\) from which we compute \(\bm{z}_{c} = \bm{z}_{\mu} + \exp(\bm{z}_{\sigma}) \cdot \bm{\epsilon} \) where elements of \(\bm{\epsilon}\) are sampled from the normal distribution.
This concludes the implementation of the probabilistic encoder $q_{\bm{\phi}}$.

\paragraph{Decoder} Recall that, in our architecture, from each latent vector representation \(\vz_{c}\) of dualized constraint \(c\) we want to compute the scalar deviation $\delta_{c}$ to the CR dual value $\lambda_{c}$ so that the sum of the two improves the Lagrangian bound given by the CR dual solution.
In other words, we want to compute \(\bm{\delta}\) such as $\vpi = [\bm{\lambda}+\bm{\delta}]_{+}$ gives a \emph{good} Lagrangian bound $LR(\bm{\pi})$. Its exact computation is of combinatorial nature and problem specific.\footnote{$LR(\vpi)$ is described in Appendices~\ref{app:MC} and~\ref{app:GA} for the two problems on which we evaluate our method.}

The probabilistic nature of the encoder-decoder can be exploited further: during evaluation, when computing a Lagrangian Relaxation, we sample constraint representations 5 times from the probabilistic encoder and return the best \(LR(\bm{\pi})\) value from the decoder.

\paragraph{Link with Energy Based Models in structured prediction}
The relaxed Lagrangian problem usually decomposes into independent subproblems due to the dualization of the linking constraints. 
In this case, for each independent Lagrangian subproblem we want to find its optimal variable assignment, usually with local combinatory constraints, for its objective reparameterized with \(\bm{\pi}\).
This approach is typical of structured prediction: we leverage neural networks to extract features in order to compute local energies (scalars), which are used by a combinatorial algorithm outputting a structure whose objective value can be interpreted as a global energy.
For instance, this is reminiscent of how graph-based syntactic parsing models in NLP compute parse scores (global energies) as sums of arc scores (local energies) computed by RNNs followed by MLPs, where the choice of arcs is guided by well-formedness constraints enforced by a maximum spanning tree solver, see for instance~\citep{kiperwasser-goldberg-2016-simple}.
Thus, the decoder is local to each dualized constraint, and we leverage subproblems to interconnect predictions:

\begin{enumerate}
	\item We compute LMs (local energies) \(\pi_{c} = [\lambda_{c} + f_{\bm{\theta}}(\bm{z}_{c})]_{+}\) for all dualized constraints $c$, where  \(f_{\bm{\theta}}\) is implemented as a feed-forward network computing the deviation.
	\item For parameter learning or if the subproblem solutions or the Lagrangian bound are the desired output, vector $\bm{\pi}$ is then passed to the Lagrangian subproblems
 which compute independently and in parallel their local solutions \(\bm{x}\) and the corresponding values are summed to give (global energy) \(LR(\bm{\pi})\).
				
\end{enumerate}

\section{ Related Work}

There is growing interest in leveraging Machine Learning (ML) alongside optimization algorithms~\citep{lodiTO2018}, in particular with the goal of improving MILP solvers's efficiency~\citep{ZHANG2023205}.
Indeed, even though MILP solvers solve problems in an exact way, they make many heuristic decisions which can be based on data-driven ML systems.
For instance, classifiers have been designed for Branch and Bound (B\&B) algorithms \citep{lodiLearningAndBranching} in order to choose which variables to branch on \citep{Alvarez2016OnlineLF,variableBranch, he-2014-learn-searc, EtheveABJK20}, which B\&B node to process \citep{Yilmaze2020,labassi2022learning}, to decide when to perform heuristics \citep{bbHeuristics, Khalil2017LearningTR} or how to schedule them \citep{chmiela2021learning}.

In this work, we depart from this main trend and predict a dual bound for MILP instances sharing common features, which can in turn be used to improve solvers.
Several propositions have tackled the prediction of high quality primal and dual bounds.
For instance, \citet{smipunn2020} predict partial variable assignments, resulting in small MILPs which can be solved to optimality.
Another way to provide primal solutions is to transform a MILP into an easier one, solve it and apply a procedure to recover primal feasibility~\citep{dalle2022learning, parmentier2022learning}.
Many works use Reinforcement Learning and  guided greedy decoding to find high-quality approximate solutions for NP-hard problems, \eg{}~\citep{kool2018attention}.
For dual bounds, ML has been employed for cut selection in cutting planes algorithms~\citep{ BalteanLugojan2018SelectingCP,wang2023learning, Balcan2021SampleCO, Berthold2022LearningTU, pmlr-v119-tang20a, HUANG2022108353, Afia2017SupervisedLI, morabit_machine-learningbased_2021}, an essential feature of MILP solvers which must balance strengthened linear relaxations with increased computations due to added cuts~\citep{dey_theoretical_2018}.

Regarding specifically prediction for Lagrangian dual solutions,  \citet{nair2018learning} consider 2-stage stochastic MILPs, approached by a Lagrangian decomposition for which they learn to predict LMs compliant with any second-stage scenario to give a good bound on average. 
\citet{pmlr-v139-lange21a} propose a heuristic to solve binary ILPs based on a specific LD where the relaxed LR problem is decomposed into many subproblems, one per constraint, solved using binary decision diagrams. This method is modified by \citet{Abbas_2022_CVPR} to be run on GPU. The block coordinate method used to heuristically solve LD is improved by learning parameters used for initializing and updating Lagrangian multipliers \citep{abbas_doge-train_2022}.
In contrast to our generic method, other previous attempts at Lagrangian dual solution prediction for deterministic MILPs focus on a specific combinatorial optimization problem, such as the cutting stock problem~\cite{initialSCG}, where a MLP predicts the dual Lagrangian value for each constraint (\emph{i.e.} stock) separately, 
or the unit commitment problem~\cite{sugishita_warmstarting_2021}, where the same problem is solved daily but with different demand forecasts with either a MLP or a random forest which predicts dual solutions used to warm-start BM.

In our work, we assume that the set of dualized constraints is given, but predicting such a set is also an active avenue of research where solutions must find a good compromise between the quality of the Lagrangian dual bound and the running time to compute this bound~\citep{KruberLP_cpaior17,basso_random_2020}.


Regarding our use of GNNs, this has become a common MILP feature extractor in recent works, either on the factor bipartite graph~\citep{gasse_exact_2019,smipunn2020} or directly on the underlying graph in routing problems~\cite{sun2023difusco}. While these works use GNNs to extract features for variable predictions, we use GNNs to extract features for constraints, which are then decoded to Lagrangian Multipliers.
Our specific GNN architecture is based on the block structure of Transformers~\cite{vaswani2017attention} where attention is replaced by a linear graph convolution.

Our method predicts a deviation from an initial solution, and can also be understood as predicting a gradient or subgradient step.
We can thus relate our approach to works on gradient descent~\citep{NIPS2016A,NEURIPS2022B} and unrolling of iterative methods for structured prediction~\citep{NIPS2016_1679091c,belanger-2017-end-end}.
This is also related to amortization especially in relation with subgradient methods already studied in the ML community~\citep{komodakis:hal-01090971,meshi2010learning}.
However, in previous works amortization was performed only during training, and iterative methods were used at testing time.
In the case of semi-amortization, our method can be used as an informed starting point for a descent algorithm applied to quadratic optimization~\cite{Sambharya2022EndtoEndLT}, or probabilistic inference~\citep{pmlr-v80-kim18e}.


\section{Evaluation}\label{sec:evaluation}

We evaluate our approach\footnote{See Appendix \ref{app:hyperparameters} for hyperparameter values used in our experiments.} on two standard problems of Operations Research, namely Multi-Commodity Fixed-Charge Network Design and  Generalized Assignment. 

\subsection{Problems and Datasets}

We review briefly the two problems and the data generation process. More details on MILP formulations and Lagrangian relaxations can be found in Appendices~\ref{app:MC} and~\ref{app:GA}
and a thorough description of dataset generation is given in Appendix~\ref{app:datasets}.

\paragraph{Multi-Commodity Fixed-Charge Network Design (MC)}
Given a network with arc capacities and a set of commodities, MC consists in activating a subset of arcs and routing each commodity from its origin to its destination, possibly fractioned on several paths, using only the activated arcs.
The objective is to minimize the total cost induced by the activation of arcs and the routing of commodities.
This problem has been used in many real-world applications for a long time, see for instance~\citep{magnanti-1984-networ-desig} for telecommunications.
It is NP-hard and its continuous relaxation provides poor bounds when arc capacities are high.
Hence, it is usually tackled with Lagrangian relaxation-based methods~\citep{mmcf}.

While the Canad dataset is the standard and well-established dataset of instances for evaluating MC solvers~\citep{crainic2001bundle}, it is too small to be used as a training set for Machine Learning where large collections of instances sharing common features are required.
Thus,
we generate new instances from a subset of instances of the Canad dataset \citep{crainic2001bundle}, that we divide in four datasets of increasing difficulty.
The first two datasets, {\small\textsc{Mc-Sml-40}} and {\small\textsc{Mc-Sml-Var}}, contain instances that all share the same network ($20$ nodes and $230$ edges) and the same arc capacities and fixed costs, but with different values for origins, destinations, volumes, and routing costs.
Instances of the former all involve the same number of commodities ($40$), while for the latter
 the number of commodities varies from $40$ to $200$.
 Dataset {\small\textsc{Mc-Big-40}} is generated similarly to {\small\textsc{Mc-Sml-40}} but upon a bigger graph containing $30$ nodes and $520$ arcs.
 Finally,  {\small\textsc{Mc-Big-Var}} contains examples generated using either the network of  {\small\textsc{Mc-Sml-40}} or the one of {\small\textsc{Mc-Big-40}}, with the number of commodities varying between $40$ and $200$.

\paragraph{Generalized Assignment (GA)}

GA consists, given a set of items and a set of capacitated bins, in assigning items to bins without exceeding their capacity in order to maximize the profit of the assignment. GA is a well-known problem in Operations Research and has numerous applications such as job-scheduling in Computer Science~\cite{balachandran_integer_1976}, distributed caching~\cite{Fleischer2011Tight} or even parking allocation~\cite{mladenovic_parking_2020}.

We generated two datasets, namely {\small\textsc{Ga-10-100}} and {\small\textsc{Ga-20-400}}, containing respectively instances with 10 bins and 100 items, and with 20 bins and 400 items. Weights, profits and bin capacities are sampled using a distribution determined from values of standard instances \cite{Yagiura1999variable}.

\subsection{Numerical Results}\label{sec:numerical}

We want to evaluate how our Lagrangian bound prediction compares to an iterative model based on subgradient, and how useful the former is as an initial point to warm-start the latter.
For that purpose, we choose a state-of-the-art proximal bundle solver provided by \textsc{Sms++}~\citep{smspp} which allows writing a MILP in a block structure fashion and using decomposition techniques to solve subproblems efficiently.
We also compare our approach with CR computed using the \textsc{Cplex}\footnote{\url{https://www.ibm.com/products/ilog-cplex-optimization-studio}} optimizer.

All MILP instances for which we want to evaluate our model are first solved by \textsc{Sms++}.
For an instance \(\iota\) we denote \(\bm{\pi}^{*}_{\iota}\) the LMs returned by \textsc{Sms++}.

\paragraph{Metrics}
We use the percentage gap as metrics to evaluate the quality of the bounds computed by the different systems, averaged over a dataset of instances \(\mathcal{I}\).
For a system returning a bound $B_\iota$ for an instance $\iota$ the percentage GAP is:

\[
  100 \times \frac{1}{| \mathcal{I} |}\sum_{\iota\in\mathcal{I}}\frac{LR(\bm{\pi}^*_{\iota})- B_{\iota} }{LR(\bm{\pi}^*_\iota)}
\]

GAP measures the quality of the bound $B_\iota$, and is zero when $B_\iota$ equals the optimal Lagrangian bound.

\paragraph{Data for Evaluation}
We divide each dataset of 2000 instances in train (80\%), validation (10\%) and test (10\%).
Parameters are learned on the train set, model selection is performed on validation set, and test proxies for unseen data.
Results are averaged over 3 random initializations.

\paragraph{Bound Accuracy}
Table~\ref{tab:tvtDataset} reports the performance of different systems on our 6 datasets.
We compare the bound returned by CR, and the bound of the Lagrangian relaxed problem obtained  with LMs computed by different methods:
\begin{itemize}
  \item LR(0) is the LR value computed with LMs set to zero.
  \item LR(CR) is the LR value computed with LMs set to CR dual solution.
  \item LR($k$-NN) is the LR value computed with LMs set to the average value of LMs from the train set returned by a $k$-NN regressor.\footnote{See Appendices~\ref{app:kNN} and \ref{app:mlpfeatures} for more information about the implemented k-NN method.}
  \item LR(MLP) is the LR value computed with LMs returned by a MLP\footnote{Additional initial features that the ones used in our model are used, see Appendix~\ref{app:mlpfeatures} for more details.} instead of the GNN-based encoder-decoder.
  \item Ours, that is the LR value computed with LMs set the output of our encoder-decoder. As written in the previous section we sample 5 LM assignments per instance and return the best LR value.
\end{itemize}

For all datasets, our method outperforms other approaches.
Our model can reach $2\%$ difference with BM on {\textsc{Mc-Sml-40}}, the easiest corpus with a small fixed network and a fixed number of commodities.
This means that one pass through our network can save numerous iterations if we can accept about $2\%$ bound error on average.
The margin with other methods is quite large for MC datasets where the CR bound is far from the optimum, with a gap reduction ranging from 77\% (\textsc{Mc-Big-Var}) to 84\% (\textsc{Mc-Sml-40}) depending on the dataset.
For GA, where CR is closer to the optimum, our model still manages to find better solutions.
Even though the gap absolute difference may seem small, the gap reduction from the second-best model LR(CR) ranges from 30\% (\textsc{Ga-10-100}) to 44\% (\textsc{Ga-20-400}), a significant error reduction.

Compared to simpler ML approaches, we see \textit{(i)} that retrieving LM values from $k$-NN clustering is not a viable solution,
even when the validation instances are close to the training instances (\textsc{Mc-Sml-40}), clustering cannot find meaningful neighbors, and \textit{(ii)} the graph feature extractor (GNN) is paramount: the LR(MLP) architecture seems unable to deviate LMs consistently from CR solutions and can even perform worse than CR or LR(CR) (\textsc{Ga-20-400}).

Regarding speed, LR(0) is the fastest since it simply amounts to solving the relaxed Lagrangian problem with the original costs.
Then CR and LR(CR) are second, the difference being that for the latter after solving CR, the dual solution $\bm{\lambda}$ is used to compute the $LR(\bm{\lambda})$.
Slowness for LR(MLP) and LR($k$-NN) is mainly caused by feature extraction (\cf{} Appendix~\ref{app:mlpfeatures}).

\begin{table}[t]
    \caption{Bound accuracies of different methods on test sets averaged by instance.}
    \centering \scriptsize
    
     \vspace{0.25cm}
    \begin{tabular}{llrr}
Dataset     & Methods   & GAP \%  & time (ms) \\
\midrule
\multirow{5}{*}{{\tiny\textsc{Mc-Sml-40}}}& CR   	 & 12.99 	 & 90.63 	 \\
& LR(0)   & 100.00  & \textbf{0.35} \\
& LR(CR)  & 12.97 	 & 90.98 	\\
& LR($k$-NN)   & 38.80 & 219.42 \\
& LR(MLP) & 10.70 & 142.48 \\
& \textbf{ours}   & \textbf{2.09} 	 &  120.96	\\
\midrule
\multirow{5}{*}{{\tiny\textsc{Mc-Sml-Var}}}& CR   	 	 & 22.29 	 & 283.63 	  \\
&  LR(0)    & 100.00  & \textbf{1.32} 	 \\
& LR(CR)    & 22.29 	 &  285.03 	  \\
& LR($k$-NN)  & 44.12 & 371.51\\
& LR(MLP) & 16.71 & 369.61\\
& \textbf{ours}     & \textbf{4.42} 	 &  374.20 	   \\
\midrule
\multirow{5}{*}{{\tiny\textsc{Mc-Big-40}}} & CR   	 	 & 15.94 	 & 220.91 	\\
& LR(0)   	 & 100.00  & \textbf{0.75} \\
& LR(CR)    & 15.85 	 & 229.57 	 \\
& LR($k$-NN)     & 54.57 & 334.99 \\
& LR(MLP) & 13.67 & 556.89 \\
& \textbf{ours}   	 & \textbf{4.20} 	  & 283.40 \\
\midrule
\multirow{5}{*}{{\tiny\textsc{Mc-Big-Var}}} & CR   	 	& 20.66 	 & 287.20 \\
&  LR(0)   & 100.00 	 &\textbf{1.37}	 \\
& LR(CR)   & 20.63 	 & 288.55 \\
& LR($k$-NN) &  49.74 & 886.91\\
& LR(MLP) & 16.14 &  515.60\\
& \textbf{ours}   	& \textbf{4.77} 	 & 374.78 \\
\midrule
\multirow{5}{*}{{\tiny\textsc{Ga-10-100}}} & CR   	   & 1.91 	 & 9.59 \\
& LR(0)   	 & 3.13	     & \textbf{ 0.44} \\
& LR(CR)    & 0.79 	 & 10.15 	 \\
& LR($k$-NN)     & 1.07 & 11.70 \\
& LR(MLP) & 0.78  & 51.71 \\
& \textbf{ours}    	 & \textbf{0.55} 	 & 16.19	\\
\midrule
\multirow{5}{*}{{\tiny\textsc{Ga-20-400}}} & CR   		  & 0.44	 &  71.40 		\\
&  LR(0)   	  & 2.70 	 & \textbf{7.51}           \\
&  LR(CR)   	  & 0.27 	 & 78.80		\\
&  LR($k$-NN) & 0.43 & 89.68 \\
& LR(MLP) &  0.28 & 114.41 \\
&  \textbf{ours}   	  & \textbf{0.15} 	 & 124.96	 \\
\bottomrule
    \end{tabular}
    \label{tab:tvtDataset}
\end{table}

\paragraph{Warm-starting Iterative Solvers}
We would like to test whether the Lagrangian Multipliers predicted by our model can be used as an informed starting point for an iterative solver for the Lagrangian Dual LD, namely the bundle method as implemented by \textsc{Sms++} and the subgradient method.
While the latter is simple to implement and only requires solving $LR(\bm{\pi})$, it has a non-smooth objective and the subgradient does not always give a descent direction, resulting in unstable updates.
In contrast, the bundle method is stabilized with a quadratic penalty assuring a smooth objective, at the expense of longer computation times. 
We hope that our model can produce good starting points for both methods and thus avoid many early iterations.

In Table~\ref{table:Bundle} we compare different initial LM vectors on the validation set of \textsc{Mc-Big-Var} for the bundle method.
We run our bundle solver until the difference between  $LR(\bm{\pi}^{*})$ and the current bound is smaller than the threshold \(\epsilon\).
We average resolution times and numbers of iterations over instances, and compute standard deviation.
We compare three initialization methods: zero, using CR dual solutions, and our model's predictions.

\begin{table*}[t]
  	\caption{Impact of initialization for a Bundle solver on \textsc{Mc-Big-Var}.
      We consider initializations from the null vector (zero), the continuous relaxation duals (CR), and our model (Ours).
}

         \vspace{0.25cm}
 \label{table:Bundle}
	\centering
  {\scriptsize
\begin{tabular}{rrrrrrrrr}
 \toprule 
		\multirow{2}{0.5cm}{$\quad \epsilon$} & \multicolumn{2}{c}{zero} & & \multicolumn{2}{c}{CR} &  & \multicolumn{2}{c}{Ours} \\
  \cmidrule{2-3} \cmidrule{5-6}\cmidrule{8-9}
	                                        & time (s)        &    \# iter.   & & time (s)    &   \# iter.      &  &     time (s)     &      \# iter.    \\
		\midrule
1e-1 & 34.34 ($\pm$81.22 ) & 90.12 ($\pm$52.41 ) & & 31.67 ($\pm$75.12 ) & 83.00 ($\pm$50.73 ) & & \textbf{16.09} ( $\pm$ 42.79 ) & \textbf{60.39} ( $\pm$ 41.37) \\

\midrule
1e-2 & 68.80 ($\pm$188.21 ) & 141.43 ($\pm$112.72 ) & & 62.09 ($\pm$171.71 ) & 133.26 ($\pm$109.60 ) & & \textbf{36.10} ( $\pm$ 106.22 ) & \textbf{105.93}( $\pm$ 97.04) \\
\midrule
1e-3 & 100.71 ($\pm$288.16 ) & 188.14 ($\pm$167.33) & & 89.26 ($\pm$251.15 ) &  179.40 ($\pm$170.15 ) & & \textbf{57.23} ($\pm$ 177.24 )  & \textbf{143.36} ($\pm$ 142.58) \\
    \midrule
1e-4 & 105.03 ($\pm$298.53 ) & 207.90 ($\pm$198.92 ) & & 101.14 ($\pm$283.52 ) &  200.42 ($\pm$197.60 ) & & \textbf{63.25} ( $\pm$ 190.47 ) & \textbf{159.42} ( $\pm$ 162.32) \\

		\bottomrule
\end{tabular}
}
\end{table*}

We can see that CR is not competitive with the null initialization, since the small gain in the number of iterations is absorbed by the supplementary computation.
However, our model's predictions give a significant improvement over the other two initialization methods, despite the additional prediction time.
Resolution time is roughly halved for the coarsest threshold, and above one-third faster for the finest one.
This is expected, as gradient-based methods naturally slow down as they approach convergence.
In appendix~\ref{app:SGDinit} we perform the same experiments as in Table~\ref{table:Bundle} for the subgradient method.

\paragraph{Ablation Study}
In Table~\ref{tab:GAPCR3models} we compare three variants of our original model, denoted \texttt{ours}, on \textsc{Mc-Sml-40} and \textsc{Mc-Big-Var}.
Results are averaged over 3 runs.

In the first variant \texttt{-max}, instead of sampling multiple LMs for each dualized constraint and keeping the best, we take one sample only per constraint. We can see that this has a minor incidence on the quality of the returned solution.
In \texttt{-sum}, the dual solution values are passed as constraint node features but are not added to the output of the decoder to produce LMs, \ie{}  the network must transport these values from its input layer to its output. 
This has a sensible negative impact of GAP scores.
In the third variant, \texttt{-cr} the CR solution is not given as input features to the network (nor added to the network's output).
This is challenging because the network does not have access to a good starting point, this is equivalent to initializing LMs to zero.
The last variant, \texttt{-sample}, uses CR as \texttt{ours} but does not sample representations \(\vz_{c}\) in the latent domain. We interpret the vector $\vh_{c}$ associated with dualized contraint $c$ after the GNN stack directly as vector $\vz_{c}$, making the encoder deterministic.

We can see that the performance of \texttt{-sum} just below \texttt{ours}, while \texttt{-cr} cannot return competitive bounds.
This indicates that the CR solution passed as input features is essential for our architecture to get good performance, whereas the computation of the deviation instead the full LM directly is not an important trait.
Still, we note that performances of $\texttt{-cr}$ should be compared with $LR(0)$ in Table~\ref{tab:tvtDataset} rather than $LR(CR)$.
In that case, we see that the GAP reduction is around $80\%$, making clear that our model is not simply repeating CR solutions.
{This means that our model could be used without the CR solution information as input, opening our methods to a wider range of problems, and paving the way for faster models.}

Finally, \texttt{-sample} is a system trained without sampling at training time, \ie{} the encoder-decoder is deterministic.
We see that sampling gives a slight performance increase on small and bigger instances.
{
In Appendix~\ref{app:LayersNumber} we compare the architecture we introduce in this work with the architecture proposed by Nair et al. in \cite{nair2018learning} and  the one presented in \cite{gasse_exact_2019}, for several layers from 1 to 10.
}

  \begin{table}[t]
     \caption{Ablation studies comparing the prediction of our model with, predicting LMs on rather than deviation from CR (\texttt{-sum}), not using CR features at all (\texttt{-cr}), or replacing the probabilistic encoder by a deterministic one (\texttt{-sample}).}\label{tab:GAPCR3models}
     \vspace{0.25cm}

  \centering
  {\scriptsize
    \begin{tabular}{r|rr}
      \toprule
    & \multicolumn{2}{c}{GAP \%} \\
        \cmidrule{2-3}
    model  	 & \textsc{Mc-Sml-40}  &	 \textsc{Mc-Big-Var} 	 	 \\
    \midrule
    \texttt{ours}                          & 2.09     &  4.77   \\
    \midrule
    \texttt{-max}                      & 2.10      &  4.79  \\
    \midrule
    \texttt{-sum}                          & 2.63     &  6.77  \\
    \texttt{-cr}                      &20.26     & 23.78  \\
    \midrule
    \texttt{-sample }                      & 2.18     &  5.86  \\

    \bottomrule
    \end{tabular}}
\end{table}

 \paragraph{Generalization Properties}

 {We test the model trained on {\small\textsc{Mc-Big-Var}} on a dataset composed of 1000 bigger instances. They are created using the biggest graph used to generate the {\small\textsc{Mc-Big-Var}} dataset but contain 160 or 200 commodities whereas the instances of {\small\textsc{Mc-Big-Var}} with the same graphs only contain up to 120 commodities.}
 In Table~\ref{tab:generalization} we can see that our model still performs well in these instances dividing by 4 the gap provided by LR(CR).
 
 \begin{table}[t]
     \caption{Generalization results over bigger instances.}
     \vspace{0.25cm}
     \label{tab:generalization}
     \centering \scriptsize
     \begin{tabular}{llrrrrr}
       \toprule
    & \multicolumn{2}{c}{GAP \%} & & \multicolumn{2}{c}{time (s)} \\
        \cmidrule{2-3} \cmidrule{5-6}
    \# commodities  	 & Ours & LR(CR) & & Ours & LR(CR) \\
         
\midrule
          160 & 6.51 & 27.85 & & 1.533 & 0.8915 \\
          200 & 7.62   &  30.18 & & 1.4328 & 1.0889        \\
          \bottomrule
     \end{tabular}
 \end{table}

\section{Conclusion}\label{sec:conclusion}

We have presented a novel method to compute good Lagrangian dual solutions for MILPs sharing common attributes, by predicting Lagrangian multipliers.
We cast this problem as an encoder-decoder prediction, where the probabilistic encoder outputs one distribution per dualized constraint from which we sample constraint vector representation.
Then a decoder transforms these representations into Lagrangian multipliers.

We experimentally showed that this method gives bounds significantly better than the commonly used heuristics on two standard combinatorial problems: it reduces the continuous relaxation gap to the optimal bound up to $85\%$, and when used to warm-start an iterative solver, the points predicted by our models reduce solving times by a large margin.

Our predictions could be exploited in primal heuristics, possibly with auxiliary losses predicting values from variable nodes, or to efficiently guide a Branch-and-Bound exact search.
Predictions could be stacked to act as an unrolled iterative solver.
Finally, we can see our model as performing denoising from a previous solution and could be adapted to fit in a diffusion model.

\section*{Acknowledgments}
The authors acknowledge the support of the French Agence Nationale de la Recherche (ANR), under grant ANR-23-CE23-0005 (project SEMIAMOR).
\section*{Impact Statements}

This paper presents work whose goal is to advance the field of Machine Learning. There are many potential societal consequences of our work, none which we feel must be specifically highlighted here.

\bibliographystyle{icml2024}
\bibliography{main}

\begin{thebibliography}{72}
\providecommand{\natexlab}[1]{#1}
\providecommand{\url}[1]{\texttt{#1}}
\expandafter\ifx\csname urlstyle\endcsname\relax
  \providecommand{\doi}[1]{doi: #1}\else
  \providecommand{\doi}{doi: \begingroup \urlstyle{rm}\Url}\fi

\bibitem[Abbas \& Swoboda(2022)Abbas and Swoboda]{Abbas_2022_CVPR}
Abbas, A. and Swoboda, P.
\newblock Fastdog: Fast discrete optimization on gpu.
\newblock In \emph{Proceedings of the IEEE/CVF Conference on Computer Vision
  and Pattern Recognition (CVPR)}, pp.\  439--449, June 2022.

\bibitem[Abbas \& Swoboda(2024)Abbas and Swoboda]{abbas_doge-train_2022}
Abbas, A. and Swoboda, P.
\newblock Doge-train: Discrete optimization on {GPU} with end-to-end training.
\newblock In Wooldridge, M.~J., Dy, J.~G., and Natarajan, S. (eds.),
  \emph{Thirty-Eighth {AAAI} Conference on Artificial Intelligence, {AAAI}
  2024, Thirty-Sixth Conference on Innovative Applications of Artificial
  Intelligence, {IAAI} 2024, Fourteenth Symposium on Educational Advances in
  Artificial Intelligence, {EAAI} 2014, February 20-27, 2024, Vancouver,
  Canada}, pp.\  20623--20631. {AAAI} Press, 2024.

\bibitem[Afia \& Kabbaj(2017, Tetouan Morocco)Afia and
  Kabbaj]{Afia2017SupervisedLI}
Afia, A.~E. and Kabbaj, M.~M.
\newblock Supervised learning in branch-and-cut strategies.
\newblock In \emph{International Conference on Big Data Cloud and
  Applications}, 2017, Tetouan Morocco.

\bibitem[{Akhavan Kazemzadeh} et~al.(2022){Akhavan Kazemzadeh}, Bektaş,
  Crainic, Frangioni, Gendron, and Gorgone]{mmcf}
{Akhavan Kazemzadeh}, M.~R., Bektaş, T., Crainic, T.~G., Frangioni, A.,
  Gendron, B., and Gorgone, E.
\newblock Node-based lagrangian relaxations for multicommodity capacitated
  fixed-charge network design.
\newblock \emph{Discrete Applied Mathematics}, 308:\penalty0 pp. 255--275,
  2022.
\newblock Combinatorial Optimization ISCO 2018.

\bibitem[Alvarez et~al.(2017)Alvarez, Louveaux, and
  Wehenkel]{Alvarez2016OnlineLF}
Alvarez, A.~M., Louveaux, Q., and Wehenkel, L.
\newblock A machine learning-based approximation of strong branching.
\newblock \emph{{INFORMS} J. Comput.}, 29\penalty0 (1):\penalty0 185--195,
  2017.

\bibitem[Andrychowicz et~al.(2016)Andrychowicz, Denil, G\'{o}mez, Hoffman,
  Pfau, Schaul, Shillingford, and de~Freitas]{NIPS2016A}
Andrychowicz, M., Denil, M., G\'{o}mez, S., Hoffman, M.~W., Pfau, D., Schaul,
  T., Shillingford, B., and de~Freitas, N.
\newblock Learning to learn by gradient descent by gradient descent.
\newblock In Lee, D., Sugiyama, M., Luxburg, U., Guyon, I., and Garnett, R.
  (eds.), \emph{Advances in Neural Information Processing Systems}, volume~29.
  Curran Associates, Inc., 2016.

\bibitem[Ba et~al.(2022)Ba, Erdogdu, Suzuki, Wang, Wu, and Yang]{NEURIPS2022B}
Ba, J., Erdogdu, M.~A., Suzuki, T., Wang, Z., Wu, D., and Yang, G.
\newblock High-dimensional asymptotics of feature learning: How one gradient
  step improves the representation.
\newblock In Koyejo, S., Mohamed, S., Agarwal, A., Belgrave, D., Cho, K., and
  Oh, A. (eds.), \emph{Advances in Neural Information Processing Systems},
  volume~35, pp.\  37932--37946. Curran Associates, Inc., 2022.

\bibitem[Ba et~al.(2016)Ba, Kiros, and Hinton]{ba2016layer}
Ba, J.~L., Kiros, J.~R., and Hinton, G.~E.
\newblock Layer normalization, 2016.

\bibitem[Balachandran(1976)]{balachandran_integer_1976}
Balachandran, V.
\newblock An {Integer} {Generalized} {Transportation} {Model} for {Optimal}
  {Job} {Assignment} in {Computer} {Networks}.
\newblock \emph{Operations Research}, 24\penalty0 (4):\penalty0 742--759,
  August 1976.

\bibitem[Balcan et~al.(2021)Balcan, Prasad, Sandholm, and
  Vitercik]{Balcan2021SampleCO}
Balcan, M.-F.~F., Prasad, S., Sandholm, T., and Vitercik, E.
\newblock Sample complexity of tree search configuration: Cutting planes and
  beyond.
\newblock \emph{Advances in Neural Information Processing Systems},
  34:\penalty0 pp. 4015--4027, 2021.

\bibitem[Baltean-Lugojan et~al.(2018)Baltean-Lugojan, Bonami, Misener, and
  Tramontani]{BalteanLugojan2018SelectingCP}
Baltean-Lugojan, R., Bonami, P., Misener, R., and Tramontani, A.
\newblock Selecting cutting planes for quadratic semidefinite
  outer-approximation via trained neural networks.
\newblock In \emph{optimization-online}, 2018.

\bibitem[Basso et~al.(2020)Basso, Ceselli, and Tettamanzi]{basso_random_2020}
Basso, S., Ceselli, A., and Tettamanzi, A.
\newblock Random sampling and machine learning to understand good
  decompositions.
\newblock \emph{Annals of Operations Research}, 284\penalty0 (2):\penalty0 pp.
  501--526, January 2020.

\bibitem[Beasley(1990)]{beasley1990lagrangian}
Beasley, J.~E.
\newblock A lagrangian heuristic for set-covering problems.
\newblock \emph{Naval Research Logistics (NRL)}, 37\penalty0 (1):\penalty0 pp.
  151--164, 1990.

\bibitem[Belanger et~al.(2017)Belanger, Yang, and
  McCallum]{belanger-2017-end-end}
Belanger, D., Yang, B., and McCallum, A.
\newblock End-to-end learning for structured prediction energy networks.
\newblock In Precup, D. and Teh, Y.~W. (eds.), \emph{Proceedings of the 34th
  International Conference on Machine Learning, {ICML} 2017, Sydney, NSW,
  Australia, 6-11 August 2017}, volume~70 of \emph{Proceedings of Machine
  Learning Research}, pp.\  429--439. {PMLR}, 2017.

\bibitem[Bengio et~al.(2021)Bengio, Lodi, and Prouvost]{lodiTO2018}
Bengio, Y., Lodi, A., and Prouvost, A.
\newblock Machine learning for combinatorial optimization: a methodological
  tour d’horizon.
\newblock \emph{European Journal of Operational Research}, 290\penalty0
  (2):\penalty0 pp. 405--421, 2021.

\bibitem[Berthold et~al.(2022)Berthold, Francobaldi, and
  Hendel]{Berthold2022LearningTU}
Berthold, T., Francobaldi, M., and Hendel, G.
\newblock Learning to use local cuts.
\newblock \emph{arXiv preprint arXiv:2206.11618}, 2022.

\bibitem[Chen et~al.(2018)Chen, Firat, Bapna, Johnson, Macherey, Foster, Jones,
  Schuster, Shazeer, Parmar, Vaswani, Uszkoreit, Kaiser, Chen, Wu, and
  Hughes]{chen-etal-2018-best}
Chen, M.~X., Firat, O., Bapna, A., Johnson, M., Macherey, W., Foster, G.,
  Jones, L., Schuster, M., Shazeer, N., Parmar, N., Vaswani, A., Uszkoreit, J.,
  Kaiser, L., Chen, Z., Wu, Y., and Hughes, M.
\newblock The best of both worlds: Combining recent advances in neural machine
  translation.
\newblock In \emph{Proceedings of the 56th Annual Meeting of the Association
  for Computational Linguistics (Volume 1: Long Papers)}, pp.\  76--86,
  Melbourne, Australia, July 2018. Association for Computational Linguistics.

\bibitem[Chmiela et~al.(2021)Chmiela, Khalil, Gleixner, Lodi, and
  Pokutta]{chmiela2021learning}
Chmiela, A., Khalil, E.~B., Gleixner, A., Lodi, A., and Pokutta, S.
\newblock Learning to schedule heuristics in branch and bound.
\newblock \emph{Advances in Neural Information Processing Systems},
  34:\penalty0 pp. 24235--24246, 2021.

\bibitem[Conforti et~al.(2014)Conforti, Cornu{\'e}jols, and
  Zambelli]{confortiIntegerProgramming2014}
Conforti, M., Cornu{\'e}jols, G., and Zambelli, G.
\newblock \emph{Integer Programming}.
\newblock {Springer}, {New York}, 2014.

\bibitem[Crainic et~al.(2001)Crainic, Frangioni, and
  Gendron]{crainic2001bundle}
Crainic, T.~G., Frangioni, A., and Gendron, B.
\newblock Bundle-based relaxation methods for multicommodity capacitated fixed
  charge network design.
\newblock \emph{Discrete Applied Mathematics}, 112\penalty0 (1-3):\penalty0 pp.
  73--99, 2001.

\bibitem[Dalle et~al.(2022)Dalle, Baty, Bouvier, and
  Parmentier]{dalle2022learning}
Dalle, G., Baty, L., Bouvier, L., and Parmentier, A.
\newblock Learning with combinatorial optimization layers: a probabilistic
  approach.
\newblock \emph{arXiv preprint arXiv:2207.13513}, 2022.

\bibitem[Dey \& Molinaro(2018)Dey and Molinaro]{dey_theoretical_2018}
Dey, S.~S. and Molinaro, M.
\newblock Theoretical challenges towards cutting-plane selection.
\newblock \emph{Mathematical Programming}, 170\penalty0 (1):\penalty0 pp.
  237--266, July 2018.
\newblock ISSN 0025-5610, 1436-4646.

\bibitem[Etheve et~al.(2020)Etheve, Al{\`{e}}s, Bissuel, Juan, and
  Kedad{-}Sidhoum]{EtheveABJK20}
Etheve, M., Al{\`{e}}s, Z., Bissuel, C., Juan, O., and Kedad{-}Sidhoum, S.
\newblock Reinforcement learning for variable selection in a branch and bound
  algorithm.
\newblock In Hebrard, E. and Musliu, N. (eds.), \emph{Integration of Constraint
  Programming, Artificial Intelligence, and Operations Research - 17th
  International Conference, {CPAIOR} 2020, Vienna, Austria, September 21-24,
  2020, Proceedings}, volume 12296 of \emph{Lecture Notes in Computer Science},
  pp.\  176--185. Springer, 2020.

\bibitem[Fleischer et~al.(2011)Fleischer, Goemans, Mirrokni, and
  Sviridenko]{Fleischer2011Tight}
Fleischer, L., Goemans, M.~X., Mirrokni, V.~S., and Sviridenko, M.
\newblock Tight approximation algorithms for maximum separable assignment
  problems.
\newblock \emph{Mathematics of Operations Research}, 36\penalty0 (3):\penalty0
  pp. 416--431, 2011.
\newblock ISSN 0364765X, 15265471.

\bibitem[Frangioni et~al.(2023)Frangioni, Iardella, and Durbano~Lobato]{smspp}
Frangioni, A., Iardella, N., and Durbano~Lobato, R.
\newblock {SMS++}, 2023.
\newblock URL \url{https://gitlab.com/smspp/smspp-project}.

\bibitem[Gasse et~al.(2019)Gasse, Chételat, Ferroni, Charlin, and
  Lodi]{gasse_exact_2019}
Gasse, M., Chételat, D., Ferroni, N., Charlin, L., and Lodi, A.
\newblock Exact {Combinatorial} {Optimization} with {Graph} {Convolutional}
  {Neural} {Networks}.
\newblock In Wallach, H., Larochelle, H., Beygelzimer, A., Alché-Buc, F.~d.,
  Fox, E., and Garnett, R. (eds.), \emph{Advances in {Neural} {Information}
  {Processing} {Systems}}, volume~32. Curran Associates, Inc., 2019.

\bibitem[Gendron et~al.(1999)Gendron, Crainic, and Frangioni]{Gendron1999}
Gendron, B., Crainic, T.~G., and Frangioni, A.
\newblock \emph{Multicommodity Capacitated Network Design}, pp.\  1--19.
\newblock Springer US, Boston, MA, 1999.

\bibitem[Geoffrion(1974)]{geoffrion_lagrangean_1974}
Geoffrion, A.~M.
\newblock Lagrangean relaxation for integer programming.
\newblock In Balinski, M.~L. (ed.), \emph{Approaches to {Integer}
  {Programming}}, pp.\  82--114. Springer Berlin Heidelberg, Berlin,
  Heidelberg, 1974.
\newblock ISBN 978-3-642-00740-8.

\bibitem[Golden et~al.(2008)Golden, Raghavan, and Wasil]{golden_vehicle_2008}
Golden, B., Raghavan, S., and Wasil, E. (eds.).
\newblock \emph{The {Vehicle} {Routing} {Problem}: {Latest} {Advances} and
  {New} {Challenges}}, volume~43 of \emph{Operations {Research}/{Computer}
  {Science} {Interfaces}}.
\newblock Springer US, Boston, MA, 2008.
\newblock ISBN 978-0-387-77777-1 978-0-387-77778-8.

\bibitem[He et~al.(2014)He, Daume~III, and Eisner]{he-2014-learn-searc}
He, H., Daume~III, H., and Eisner, J.~M.
\newblock Learning to search in branch and bound algorithms.
\newblock In Ghahramani, Z., Welling, M., Cortes, C., Lawrence, N., and
  Weinberger, K. (eds.), \emph{Advances in Neural Information Processing
  Systems}, volume~27. Curran Associates, Inc., 2014.

\bibitem[He et~al.(2016)He, Zhang, Ren, and Sun]{he2016residual}
He, K., Zhang, X., Ren, S., and Sun, J.
\newblock {Deep Residual Learning for Image Recognition}.
\newblock In \emph{Proceedings of 2016 IEEE Conference on Computer Vision and
  Pattern Recognition}, CVPR '16, pp.\  770--778. IEEE, June 2016.

\bibitem[Hiriart-Urruty \& Lemar{\'e}chal(1996)Hiriart-Urruty and
  Lemar{\'e}chal]{hiriart1996convex2}
Hiriart-Urruty, J.-B. and Lemar{\'e}chal, C.
\newblock \emph{Convex analysis and minimization algorithms II: Advance Theory
  and Bundle Methods}, volume 305.
\newblock Springer science \& business media, 1996.

\bibitem[Hottung et~al.(2020)Hottung, Tanaka, and Tierney]{bbHeuristics}
Hottung, A., Tanaka, S., and Tierney, K.
\newblock Deep learning assisted heuristic tree search for the container
  pre-marshalling problem.
\newblock \emph{Computers \& Operations Research}, 113:\penalty0 104781, 2020.

\bibitem[Huang et~al.(2022)Huang, Wang, Liu, Zhen, Zhang, Yuan, Hao, Yu, and
  Wang]{HUANG2022108353}
Huang, Z., Wang, K., Liu, F., Zhen, H.-L., Zhang, W., Yuan, M., Hao, J., Yu,
  Y., and Wang, J.
\newblock Learning to select cuts for efficient mixed-integer programming.
\newblock \emph{Pattern Recognition}, 123:\penalty0 108353, 2022.
\newblock ISSN 0031-3203.

\bibitem[Khalil et~al.(2016)Khalil, Le~Bodic, Song, Nemhauser, and
  Dilkina]{variableBranch}
Khalil, E., Le~Bodic, P., Song, L., Nemhauser, G., and Dilkina, B.
\newblock Learning to branch in mixed integer programming.
\newblock \emph{Proceedings of the AAAI Conference on Artificial Intelligence},
  30\penalty0 (1), Feb. 2016.

\bibitem[Khalil et~al.(2017)Khalil, Dilkina, Nemhauser, Ahmed, and
  Shao]{Khalil2017LearningTR}
Khalil, E.~B., Dilkina, B., Nemhauser, G.~L., Ahmed, S., and Shao, Y.
\newblock Learning to run heuristics in tree search.
\newblock In \emph{Ijcai}, pp.\  659--666, 2017.

\bibitem[Kim et~al.(2018)Kim, Wiseman, Miller, Sontag, and
  Rush]{pmlr-v80-kim18e}
Kim, Y., Wiseman, S., Miller, A., Sontag, D., and Rush, A.
\newblock Semi-amortized variational autoencoders.
\newblock In Dy, J. and Krause, A. (eds.), \emph{Proceedings of the 35th
  International Conference on Machine Learning}, volume~80 of \emph{Proceedings
  of Machine Learning Research}, pp.\  2678--2687. PMLR, 10--15 Jul 2018.

\bibitem[Kingma \& Welling(2014)Kingma and Welling]{Kingma2014}
Kingma, D.~P. and Welling, M.
\newblock Auto-encoding variational bayes.
\newblock In Bengio, Y. and LeCun, Y. (eds.), \emph{2nd International
  Conference on Learning Representations, {ICLR} 2014, Banff, AB, Canada, April
  14-16, 2014, Conference Track Proceedings}, 2014.

\bibitem[Kiperwasser \& Goldberg(2016)Kiperwasser and
  Goldberg]{kiperwasser-goldberg-2016-simple}
Kiperwasser, E. and Goldberg, Y.
\newblock Simple and accurate dependency parsing using bidirectional {LSTM}
  feature representations.
\newblock \emph{Transactions of the Association for Computational Linguistics},
  4:\penalty0 pp. 313--327, 2016.

\bibitem[Kipf \& Welling(2017)Kipf and Welling]{kipf2016semi}
Kipf, T. and Welling, M.
\newblock Semi-supervised classification with graph convolutional networks.
\newblock In \emph{International Conference on Learning Representations}, 2017.

\bibitem[Komodakis et~al.(2014)Komodakis, Xiang, and
  Paragios]{komodakis:hal-01090971}
Komodakis, N., Xiang, B., and Paragios, N.
\newblock {A Framework for Efficient Structured Max-Margin Learning of
  High-Order MRF Models}.
\newblock Technical Report~7, 2014.

\bibitem[Kool et~al.(2019)Kool, van Hoof, and Welling]{kool2018attention}
Kool, W., van Hoof, H., and Welling, M.
\newblock Attention, learn to solve routing problems!
\newblock In \emph{International Conference on Learning Representations}, 2019.

\bibitem[Korte \& Vygen(2012)Korte and
  Vygen]{korteCombinatorialOptimizationTheory2012}
Korte, B.~H. and Vygen, J.
\newblock \emph{Combinatorial Optimization: Theory and Algorithms}.
\newblock Number v. 21 in Algorithms and Combinatorics. {Springer}, {Heidelberg
  ; New York}, 5th ed edition, 2012.

\bibitem[Kraul et~al.(2023)Kraul, Seizinger, and Brunner]{initialSCG}
Kraul, S., Seizinger, M., and Brunner, J.~O.
\newblock Machine learning–supported prediction of dual variables for the
  cutting stock problem with an application in stabilized column generation.
\newblock \emph{INFORMS Journal on Computing}, 35\penalty0 (3):\penalty0 pp.
  692--709, 2023.

\bibitem[Kruber et~al.(2017)Kruber, L{\"{u}}bbecke, and
  Parmentier]{KruberLP_cpaior17}
Kruber, M., L{\"{u}}bbecke, M.~E., and Parmentier, A.
\newblock Learning when to use a decomposition.
\newblock In Salvagnin, D. and Lombardi, M. (eds.), \emph{Integration of {AI}
  and {OR} Techniques in Constraint Programming - 14th International
  Conference, {CPAIOR} 2017, Padua, Italy, June 5-8, 2017, Proceedings}, volume
  10335 of \emph{Lecture Notes in Computer Science}, pp.\  202--210. Springer,
  2017.

\bibitem[Labassi et~al.(2022)Labassi, Ch{\'e}telat, and
  Lodi]{labassi2022learning}
Labassi, A.~G., Ch{\'e}telat, D., and Lodi, A.
\newblock Learning to compare nodes in branch and bound with graph neural
  networks.
\newblock \emph{Advances in Neural Information Processing Systems},
  35:\penalty0 pp. 32000--32010, 2022.

\bibitem[Lange \& Swoboda(2021)Lange and Swoboda]{pmlr-v139-lange21a}
Lange, J.-H. and Swoboda, P.
\newblock Efficient message passing for 0{–}1 ilps with binary decision
  diagrams.
\newblock In Meila, M. and Zhang, T. (eds.), \emph{Proceedings of the 38th
  International Conference on Machine Learning}, volume 139 of
  \emph{Proceedings of Machine Learning Research}, pp.\  pp. 6000--6010. PMLR,
  18--24 Jul 2021.

\bibitem[Le et~al.(2007)Le, Smola, and Vishwanathan]{le2007bundle}
Le, Q., Smola, A., and Vishwanathan, S.
\newblock Bundle methods for machine learning.
\newblock \emph{Advances in neural information processing systems}, 20, 2007.

\bibitem[Le~Cun et~al.(2006)Le~Cun, Chopra, Hadsell, Ranzato, and
  Huang]{lecun2006tutorial}
Le~Cun, Y., Chopra, S., Hadsell, R., Ranzato, M., and Huang, F.
\newblock A tutorial on energy-based learning.
\newblock \emph{Predicting structured data}, 1\penalty0 (0), 2006.

\bibitem[Lodi \& Zarpellon(2017)Lodi and Zarpellon]{lodiLearningAndBranching}
Lodi, A. and Zarpellon, G.
\newblock On learning and branching: a survey.
\newblock \emph{Top}, 25:\penalty0 pp. 207--236, 2017.

\bibitem[Magnanti \& Wong(1984)Magnanti and Wong]{magnanti-1984-networ-desig}
Magnanti, T.~L. and Wong, R.~T.
\newblock Network design and transportation planning: Models and algorithms.
\newblock \emph{Transp. Sci.}, 18\penalty0 (1):\penalty0 pp. 1--55, 1984.

\bibitem[Meshi et~al.(2010)Meshi, Sontag, Jaakkola, and
  Globerson]{meshi2010learning}
Meshi, O., Sontag, D., Jaakkola, T., and Globerson, A.
\newblock Learning efficiently with approximate inference via dual losses.
\newblock In \emph{Proceedings of the 27th International Conference on
  International Conference on Machine Learning}, ICML'10, pp.\  783–790,
  Madison, WI, USA, 2010. Omnipress.
\newblock ISBN 9781605589077.

\bibitem[Mladenović et~al.(2020)Mladenović, Delot, Laporte, and
  Wilbaut]{mladenovic_parking_2020}
Mladenović, M., Delot, T., Laporte, G., and Wilbaut, C.
\newblock The parking allocation problem for connected vehicles.
\newblock \emph{Journal of Heuristics}, 26\penalty0 (3):\penalty0 377--399,
  June 2020.
\newblock ISSN 1381-1231, 1572-9397.

\bibitem[Morabit et~al.(2021)Morabit, Desaulniers, and
  Lodi]{morabit_machine-learningbased_2021}
Morabit, M., Desaulniers, G., and Lodi, A.
\newblock Machine-{Learning}–{Based} {Column} {Selection} for {Column}
  {Generation}.
\newblock \emph{Transportation Science}, 55\penalty0 (4):\penalty0 815--831,
  July 2021.
\newblock ISSN 0041-1655, 1526-5447.

\bibitem[Nair et~al.(2018)Nair, Dvijotham, Dunning, and
  Vinyals]{nair2018learning}
Nair, V., Dvijotham, D., Dunning, I., and Vinyals, O.
\newblock Learning fast optimizers for contextual stochastic integer programs.
\newblock In \emph{UAI}, pp.\  591--600, 2018.

\bibitem[Nair et~al.(2020)Nair, Bartunov, Gimeno, von Glehn, Lichocki, Lobov,
  O'Donoghue, Sonnerat, Tjandraatmadja, Wang, Addanki, Hapuarachchi, Keck,
  Keeling, Kohli, Ktena, Li, Vinyals, and Zwols]{smipunn2020}
Nair, V., Bartunov, S., Gimeno, F., von Glehn, I., Lichocki, P., Lobov, I.,
  O'Donoghue, B., Sonnerat, N., Tjandraatmadja, C., Wang, P., Addanki, R.,
  Hapuarachchi, T., Keck, T., Keeling, J., Kohli, P., Ktena, I., Li, Y.,
  Vinyals, O., and Zwols, Y.
\newblock Solving mixed integer programs using neural networks.
\newblock \emph{CoRR}, abs/2012.13349, 2020.

\bibitem[Parmentier(2022)]{parmentier2022learning}
Parmentier, A.
\newblock Learning to approximate industrial problems by operations research
  classic problems.
\newblock \emph{Oper. Res.}, 70\penalty0 (1):\penalty0 606--623, 2022.

\bibitem[Polyak(1987)]{Polyak87}
Polyak, B.
\newblock \emph{Introduction to Optimization}.
\newblock Optimization Software, New York, 1987.

\bibitem[Rybkin et~al.(2021)Rybkin, Daniilidis, and
  Levine]{pmlr-v139-rybkin21a}
Rybkin, O., Daniilidis, K., and Levine, S.
\newblock Simple and effective vae training with calibrated decoders.
\newblock In Meila, M. and Zhang, T. (eds.), \emph{Proceedings of the 38th
  International Conference on Machine Learning}, volume 139 of
  \emph{Proceedings of Machine Learning Research}, pp.\  9179--9189. PMLR,
  18--24 Jul 2021.

\bibitem[Sambharya et~al.(2022)Sambharya, Hall, Amos, and
  Stellato]{Sambharya2022EndtoEndLT}
Sambharya, R., Hall, G., Amos, B., and Stellato, B.
\newblock End-to-end learning to warm-start for real-time quadratic
  optimization.
\newblock In \emph{Conference on Learning for Dynamics \& Control}, 2022.

\bibitem[Schulman et~al.(2015)Schulman, Heess, Weber, and
  Abbeel]{NIPS2015_de03beff}
Schulman, J., Heess, N., Weber, T., and Abbeel, P.
\newblock Gradient estimation using stochastic computation graphs.
\newblock In Cortes, C., Lawrence, N., Lee, D., Sugiyama, M., and Garnett, R.
  (eds.), \emph{Advances in Neural Information Processing Systems}, volume~28.
  Curran Associates, Inc., 2015.

\bibitem[Srivastava et~al.(2014)Srivastava, Hinton, Krizhevsky, Sutskever, and
  Salakhutdinov]{srivastava2014dropout}
Srivastava, N., Hinton, G., Krizhevsky, A., Sutskever, I., and Salakhutdinov,
  R.
\newblock Dropout: a simple way to prevent neural networks from overfitting.
\newblock \emph{The journal of machine learning research}, 15\penalty0
  (1):\penalty0 pp. 1929--1958, 2014.

\bibitem[Sugishita et~al.(2024)Sugishita, Grothey, and
  McKinnon]{sugishita_warmstarting_2021}
Sugishita, N., Grothey, A., and McKinnon, K.
\newblock Use of {Machine} {Learning} {Models} to {Warmstart} {Column}
  {Generation} for {Unit} {Commitment}.
\newblock \emph{INFORMS Journal on Computing}, January 2024.
\newblock ISSN 1091-9856, 1526-5528.

\bibitem[Sun \& Yang(2023)Sun and Yang]{sun2023difusco}
Sun, Z. and Yang, Y.
\newblock {DIFUSCO}: Graph-based diffusion solvers for combinatorial
  optimization.
\newblock In \emph{Thirty-seventh Conference on Neural Information Processing
  Systems}, 2023.

\bibitem[Tang et~al.(2020)Tang, Agrawal, and Faenza]{pmlr-v119-tang20a}
Tang, Y., Agrawal, S., and Faenza, Y.
\newblock Reinforcement learning for integer programming: Learning to cut.
\newblock In III, H.~D. and Singh, A. (eds.), \emph{Proceedings of the 37th
  International Conference on Machine Learning}, volume 119 of
  \emph{Proceedings of Machine Learning Research}, pp.\  9367--9376. PMLR,
  13--18 Jul 2020.

\bibitem[Vaswani et~al.(2017)Vaswani, Shazeer, Parmar, Uszkoreit, Jones, Gomez,
  Kaiser, and Polosukhin]{vaswani2017attention}
Vaswani, A., Shazeer, N., Parmar, N., Uszkoreit, J., Jones, L., Gomez, A.~N.,
  Kaiser, {\L}., and Polosukhin, I.
\newblock Attention is all you need.
\newblock \emph{Advances in neural information processing systems}, 30, 2017.

\bibitem[Wang et~al.(2023)Wang, Li, Wang, Kuang, Yuan, Zeng, Zhang, and
  Wu]{wang2023learning}
Wang, Z., Li, X., Wang, J., Kuang, Y., Yuan, M., Zeng, J., Zhang, Y., and Wu,
  F.
\newblock Learning cut selection for mixed-integer linear programming via
  hierarchical sequence model.
\newblock In \emph{The Eleventh International Conference on Learning
  Representations}, 2023.

\bibitem[Wolsey(2021)]{wolseyIntegerProgramming2021}
Wolsey, L.~A.
\newblock \emph{Integer Programming}.
\newblock {Wiley}, {Hoboken, NJ}, second edition edition, 2021.

\bibitem[Yagiura et~al.(1999)Yagiura, Yamaguchi, and
  Ibaraki]{Yagiura1999variable}
Yagiura, M., Yamaguchi, T., and Ibaraki, T.
\newblock \emph{A Variable Depth Search Algorithm for the Generalized
  Assignment Problem}, pp.\  459--471.
\newblock Springer US, Boston, MA, 1999.
\newblock ISBN 978-1-4615-5775-3.

\bibitem[Yang et~al.(2016)Yang, Sun, Li, and Xu]{NIPS2016_1679091c}
Yang, Y., Sun, J., Li, H., and Xu, Z.
\newblock Deep admm-net for compressive sensing mri.
\newblock In Lee, D., Sugiyama, M., Luxburg, U., Guyon, I., and Garnett, R.
  (eds.), \emph{Advances in Neural Information Processing Systems}, volume~29.
  Curran Associates, Inc., 2016.

\bibitem[Yilmaz \& Yorke-Smith(2021)Yilmaz and Yorke-Smith]{Yilmaze2020}
Yilmaz, K. and Yorke-Smith, N.
\newblock A study of learning search approximation in mixed integer branch and
  bound: Node selection in scip.
\newblock \emph{AI}, 2\penalty0 (2):\penalty0 pp. 150--178, 2021.
\newblock ISSN 2673-2688.

\bibitem[Zhang et~al.(2023)Zhang, Liu, Li, Zhen, Yuan, Li, and
  Yan]{ZHANG2023205}
Zhang, J., Liu, C., Li, X., Zhen, H.-L., Yuan, M., Li, Y., and Yan, J.
\newblock A survey for solving mixed integer programming via machine learning.
\newblock \emph{Neurocomputing}, 519:\penalty0 pp. 205--217, 2023.

\end{thebibliography}

\appendix

\newpage

\section{Initial Features}
\label{sec:initial-features}

To extract useful features, we define a network based on graph convolutions presented in Figure~\ref{fig:model} in the line of the work of \citep{gasse_exact_2019} on MILP encoding. We detail the initial node features \({\{\bm{e}_{n}\}}_{n}\) of the MILP-encoding bipartite graph presented in Section~\ref{sec:enc-dec}.

Given an instance of the form:
\begin{subequations}\label{eq:InitFeatures_MILP}
\begin{align}
(P) \qquad & \min_{\bm{x}} \vw^{\top}\vx \\
\label{InitFeatures:ctn} & \mA\vx \begin{pmatrix} \ge \\ =\end{pmatrix}\vb  \\
& \vx \in \mathbb{R}^m_+\times\mathbb{N}^p
\end{align}
\end{subequations}
we consider the following initial features for a variable \(x_{j}\):
\begin{itemize}
	\item its coefficient \(w_{j}\) in the objective function;
	\item its value in the primal solution of CR;
	\item its reduced cost \(\bar{c}_j = w_j - \bm{\lambda}^T \mA_{j}\) in $CR$ where $\mA_j$ is the $j^{th}$ column of $\mA$ and $\bm{\lambda}$ is the dual solution of $CR$;
	\item a binary value indicating whether $x_{j}$ is integral or continuous.
\end{itemize}

For constraint $\bm{a}^\top \vx \begin{pmatrix} \ge \\ =\end{pmatrix} b$ of \eqref{InitFeatures:ctn}, we consider:
\begin{itemize}
	\item the right-hand side $b$ of the constraint;
	\item the value of the associated dual solution in $CR$;
	\item one binary value indicating whether the constraint is an equality or an inequality;
	\item one binary value stating whether $c$ is dualized in the relaxed Lagrangian problem.
\end{itemize}

We use for each node $n$ of the bipartite graph a feature vector $\bm{e}_n \in \mathbb{R}^{8}$. The first four components are used to encode the initial features if $n$ corresponds to a variable and are set to 0 otherwise, whereas the next four components are used only if $n$ is associated with a constraint and are set to 0 otherwise.

\section{Multi Commodity Capacitated Network Design Problem \label{app:MC}}

A MC instance is given by a directed simple graph $D = (N, A)$, a set of commodities $K$, an arc-capacity vector $c$, and two cost vectors $r$ and $f$.
Each commodity $k\in K$ corresponds to a triplet $(o^k, d^k, q^k)$ where $o^k\in N$ and $d^k\in N$ are the nodes corresponding to the origin and the destination of commodity $k$, and $q^k \in \mathbb{N}^*$ is its volume. For each arc, $(i,j) \in A$, $c_{ij}>0$ corresponds to the maximum amount of flow that can be routed through $(i,j)$ and $f_{ij} > 0$ corresponds to the fixed cost of using arc $(i,j)$ to route commodities. For each arc $(i,j) \in A$ and each commodity $k \in K$, $r_{ij}^k >0$
corresponds to the cost of routing one unit of commodity $k$ through arc $(i,j)$.

A MC solution consists of an arc subset $A' \subseteq A$ and, for each commodity $k \in K$, in a flow of value $q^k$ from its origin $o^k$ to its destination $d^k$ with the following requirements: all commodities are only routed through arcs of $A'$, and the total amount of flow routed through each arc $(i,j) \in A'$ does not exceed its capacity $c_{ij}$. The solution cost is the sum of the fixed costs over the arcs of $A'$ plus the routing cost, the latter being the sum over all arcs $(i,j) \in A$ and all commodities $k \in K$ of the unitary routing cost $r_{ij}^k$ multiplied by the amount of flow of $k$ routed through $(i,j)$.

\subsection{MILP formulation}


A standard model for the MC problem~\citep{Gendron1999} introduces two sets of variables: the continuous flow variables $x_{ij}^k$ representing the amount of commodity $k$ that is routed through arc $(i,j)$ and the binary design variables $y_{ij}$ representing whether or not arc $(i,j)$ is used to route commodities.
Denoting respectively by $N^+_i=\{j \in N \mid (i,j)\in A\}$ and $N_i^-=\{j \in N \mid (j, i)\in A\}$ the sets of forward and backward neighbors of a vertex $i \in N$, the MC problem can be modeled as follows:
\begin{subequations}\label{eq:MC_MILP}
\begin{align}
	\label{znd:obj}
	&\min_{\bm{x}, \bm{y}}   \sum_{(i,j) \in A} \left(f_{ij}y_{ij} + \sum_{k \in K} r^k_{ij}x_{ij}^k\right)  \span\span                                       \\
	\label{znd:flowCons}  & \sum_{j\in N^+_i}x_{ij}^k - \sum_{j\in N^-_i}x_{ji}^k=b^k_i & & \!\!\forall i \in N, \forall k \in K\\
	\label{znd:capacity}      & \sum_{k \in K}x_{ij}^k\leq c_{ij}y_{ij},            &  &\!\! \forall (i,j)\in A                  \\
    \label{znd:xij0}  &  x_{ij}^k = 0 && \!\!\!\!\!\begin{array}{l}\forall k \in K, \forall (i,j) \in A  \\ \text{ s.t. }i=d^k \text{ or } j=o^k\end{array}\\
	\label{znd:bound}         & 0\leq x^k_{ij}\leq q^k                              &  & \!\!\forall (i,j)\in A, \forall k \in K \\
	\label{znd:binary}        & y_{ij} \in \{0,1\},                              &     & \!\!\forall (i,j)\in A
\end{align}
\end{subequations}
where
\begin{equation*}
	b^k_i=\left\{ \begin{array}{cc}
		q^k  & \mbox{ if } i = o^k, \\
		-q^k & \mbox{ if } i = d^k, \\
		0    & \mbox{ otherwise.}
	\end{array}\right.
\end{equation*}

The objective function \eqref{znd:obj} minimizes the sum of the routing and fixed costs.  Equations \eqref{znd:flowCons} are the flow conservation constraints that properly define the flow of each commodity through the graph.  Constraints \eqref{znd:capacity} are the capacity constraints ensuring that the total amount of flow routed through each arc does not exceed its capacity or is zero if the arc is not used to route commodities. Equations~\eqref{znd:xij0} ensure that a commodity is not routed on an arc entering its origin or leaving its destination. 
Finally inequalities \eqref{znd:bound} are the bounds for the $x$ variables and  inequalities \eqref{znd:binary} are the integer constraints for the design variables.


\subsection{Lagrangian Knapsack Relaxation}

A standard way to obtain good bounds for the MC problem is to solve the Lagrangian relaxation obtained by dualizing the flow conservation constraints \eqref{znd:flowCons} in formulation  \eqref{znd:obj}-\eqref{znd:binary}.
Let $\pi_i^k$ be the Lagrangian multiplier associated with node $i \in N$ and commodity $k \in K$.
Dualizing the flow conservation constraints gives the following relaxed Lagrangian problem $LR(\bm{\pi})$\footnote{Since the dualized constraints are equations, $\vpi$ have no sign constraints.}:
$$ \everymath={\displaystyle}
	\begin{array}{l} \min_{(\bm{x},\bm{y}) \text{ satisfies }\eqref{znd:capacity}-\eqref{znd:binary}}    \sum_{(i,j) \in A} \left(f_{ij}y_{ij} + \sum_{k \in K} r^k_{ij}x_{ij}^k\right)\\+\sum_{k \in K}\sum_{i \in N}\pi^k_i\left(b^k_i -\sum_{j\in N^+_i}x_{ij}^k + \sum_{j\in N^-_i}x_{ji}^k\right)\end{array}
$$
Rearranging the terms in the objective function and observing that the relaxed Lagrangian problem is decomposed by arcs, we obtain a subproblem for each arc $(i,j)\in A$ of the form:
\begin{subequations}\label{eq:LRij_MC}
\begin{align}
(LR_{ij}(\bm{\pi}))\quad & \min_{\bm{x}, \bm{y}} f_{ij}y_{ij} + \sum_{k \in K_{ij}} w_{ij}^k x_{ij}^k  \span\span                       \\
& \sum_{k \in K_{ij}}x_{ij}^k\leq c_{ij}y_{ij} \label{LR_MC:capacity}      \\
& 0\leq x^k_{ij}\leq q^k & \forall k \in K_{ij} \label{LR:MC:xBounds}\\
& y_{ij} \in \{0,1\}     & \label{LR_MC:yBinary}
\end{align}
\end{subequations}
where $w_{ij}^k = r_{ij}^k-\pi_{i}^k+\pi_{j}^k$ and $K_{ij} = \{k \in K \mid j \neq o^k \mbox{ and } i \neq d^k\}$ is the set of commodities that may be routed through arc $(i,j)$.

For each $(i,j) \in A$, $LR_{ij}(\bm{\pi})$ is a MILP with only one binary variable. If $y_{ij} = 0$, then, by \eqref{LR_MC:capacity} and \eqref{LR:MC:xBounds}, $x_{ij}^k = 0$ for all $k \in K_{ij}$. If $y_{ij} = 1$, the problem reduces to a continuous knapsack problem. An optimal solution is obtained by ordering the commodities of $K_{ij}$ with respect to decreasing values $w_{ij}^k$ and setting for each variable $x_{ij}^k$ the value $\max\{\min\{q^k, c_{ij} - \sum_{k \in K(k)} q^k\}, 0\}$ where $K(k)$ denotes the set of commodities that preceed $k$ in the order. This step can be done in $O(|K_{ij}|)$ if one computes $x_{ij}^k$ following the computed order. Hence, the complexity of the continuous knapsack problem is $O(|K_{ij}|\log(|K_{ij}|))$. The solution of $LR_{ij}(\bm{\pi})$ is the minimum between the cost of the continuous knapsack problem and $\mathbf{0}$.


Lagrangian duality implies that 
\[LR(\bm{\pi})=\sum_{(i,j) \in A} LR_{ij}(\bm{\pi})+ \sum_{i \in N}\sum_{k \in K} \pi_i^k b^k_i
\] 
is a lower bound for the MC problem and the best one is obtained by solving the following Lagrangian dual problem:
\begin{equation*}
	(LD) \quad  \max_{\bm{\pi} \in \mathbb{R}^{N \times K}} LR(\bm{\pi})
\end{equation*}

\section{Generalized Assignment Problem}
\label{app:GA}

A GA instance is defined by a set $I$ of items and a set $J$ of bins.
Each bin $j$ is associated with a certain capacity $c_j$.
For each item $i \in I$ and each bin $j \in J$, $p_{ij}$ is the profit of assigning item $i$ to bin $j$, and $w_{ij}$ is the weight of item $i$ inside bin $j$.

Considering a binary variable $x_{ij}$ for each item and each bin that is equal to one if and only if item $i$ is assigned to bin $j$, the GA problem can be formulated as:
\begin{subequations}\label{eq:GA_MILP}
\begin{align}
& \max_{\bm{x}}  \sum_{i \in I} \sum_{j \in J} p_{ij}x_{ij} \label{obj:GA}\\
& \sum_{j \in J} x_{ij}  \leq 1 && \forall i \in I \label{sa_constr:GA} \\
& \sum_{i \in I} w_{ij} x_{ij}  \leq c_j && \forall j \in J \label{cap_constr:GA}\\
& x_{ij} \in \{0,1\} && \forall i \in I ,\; \forall j\in J .\label{int_constr:GA}
\end{align}
\end{subequations}

The objective function \eqref{obj:GA} maximizes the total profit.
Inequalities \eqref{sa_constr:GA} assert that each item is contained in no more than one bin.
Inequalities \eqref{cap_constr:GA} ensure that the sum of the weights of the items assigned to a bin does not exceed its capacity.
Finally, constraints \eqref{int_constr:GA} assure the integrality of the variables.

\subsection{Lagrangian Relaxation}

A Lagrangian relaxation of the GA problem is obtained by dualizing \eqref{sa_constr:GA}.
For $i \in I$, let $\pi_{i} \geq 0$ be the Lagrangian multiplier of inequality \eqref{sa_constr:GA} associated with item $i$.
For each bin $j$ the subproblem becomes:
\begin{align*}
(LR_j(\bm{\pi})) \quad & \max_{\bm{x}} \sum_{i \in I}\sum_{j \in J} (p_{ij}-\pi_{i}) x_{ij} \span \span\\
& \sum_{i \in I}w_{ij}x_{ij} \leq c_j\\
& x_{ij} \in \{0,1\} && \forall i \in I
\end{align*}
It corresponds to an integer knapsack with $|I|$ binary variables. For $\bm{\pi}\ge \bm{0}$, the Lagrangian bound $LR(\bm{\pi})$ is:
\begin{equation*}
    LR(\bm{\pi}) = \sum_{j \in J}LR_j(\bm{\pi})+\sum_{i \in I}\pi_i.
\end{equation*}

The Lagrangian dual can be then written as:
\begin{equation*}
    \min_{\bm{\pi} \in \R^{|I|}_{\geq 0}}LR(\bm{\pi})
\end{equation*}

\section{Dataset collection details \label{app:datasets}}

In this appendix we provide further details on the dataset construction.

\paragraph{Multi-Commodity Fixed-Charge Network Design}

We generate four datasets of 2000 instances each (1600 for training, 200 for validation and 200 for test) based on Canad instances \citep{crainic2001bundle}. These canad instances have been chosen such that the Lagrangian dual bound can be solved in nearly one second for the easiest instances and in approximately one hour for the hardest ones.

The first two datasets {\small\textsc{Mc-Sml-40}} and {\small\textsc{Mc-Sml-Var}} consider the same graph with 20 nodes and 230 edges, and the same capacity and fixed cost vectors. The first dataset has only instances with 40 commodities whereas the second one has instances with 40, 80, 120, 160 or 200 commodities.

Origins and destinations are randomly chosen using a uniform distribution. Volumes and routing costs are randomly sampled using a Gaussian distribution.  Sampling uses four different means $\mu$ and variances $\sigma^2$ which are determined from the four canad instances p33, p34, p35 and p36 (having the same graph and fixed costs as the datasets) in order to generate four different types of instances: whether the fixed costs are high with respect to routing costs, and whether capacities are high with respect to commodity volumes.


The third dataset {\small\textsc{Mc-Big-40}} is generated similarly as the first one except that it based on a graph with 30 nodes and 520 edges. The means and variances used to sample the fixed costs and the volumes are determined from the four canad instances p49, p50, p51 and p52. The number of commodities is equal to 40 in each instance.

Finally, the last dataset {\small\textsc{Mc-Big-Var}} contains instances with either the graph, capacities and fixed costs of the first two datasets or the ones of the third dataset. Sampling uses either the canad instances p33, p34, p35 and p36 or the canad instances p49, p50, p51 and p52 for determining the mean and variance, depending on the size of the graph. The number of commodities varies from 40 to 200 if the graph is the one of the first two datasets, and from 40 to 120 otherwise.

\paragraph{Generalized Assignment}

We create two datasets of GA instances containing 2000 instances each (1600 for training, 200 for validation and 200 for test). The first one contains instances with 10 bins and 100 items whereas the second one contains instances with 20 bins and 400 items. For each dataset, all instances are generated by randomly sampling capacities, weights and profits using a Gaussian distribution of mean $\mu$ and variance $\sigma^2$ and the values are clipped to an interval \([a,b]\). The values $\mu$, $\sigma^2$, \(a\) and \(b\) are determined from the instance e10100 for the first dataset, and from the instance e20400 for the second one \footnote{{Instances e10100 and e20400 are GA instances generated by \cite{Yagiura1999variable} and available at \url{http://www.al.cm.is.nagoya-u.ac.jp/~yagiura/gap/}.}}. More specifically, for each type of data (capacities, weights and profits), $\mu$ and $\sigma^2$ are given by the average and variance of the values of the instance, and \(a\) and \(b\) are fixed to 0.8 times the minimum value and 1.2 times the maximum value, respectively.



\section{Hyperparameters \label{app:hyperparameters}}

\paragraph{Model Architecture}

For all datasets, the MLP $F$ from initial features to high-dimensional is implemented as a linear transformation  ($8$ to $250$) followed by a non-linear activation.
Then, we consider a linear transformation to the size of the internal representation of nodes for the GNN.

For MC we use 5 blocks, while for GA we use only 3.
The fact that for GA are sufficient fewer layers can be explained by looking at the bipartite-graph representation of the instance that is denser for GA than for MC. For instance, in MC, a variable $x_{ij}^k$ appears in three constraints involving several variables while in GA, each variable $x_{ij}$ appears in $|I| + |J|$ constraints so the propagation needs fewer convolutions for the information to be propagated.

The hidden layer of the MLP in the second sub-layer of each block has a size of 1000.

The decoder is an MLP with one hidden layer of $250$ nodes.

All non-linear activations are implemented as ReLU.
Only the one for the output of the GA is a softplus.

The dropout rate is set to 0.25.

\paragraph{Optimiser Specifications}
We use as optimizer RAdam, with learning rate $0.0001$ for MC and $0.00001$ for GA, a Clip Norm (to 5) and exponential decay $0.9$, step size $100000$ and minimum learning rate $10^{-10}$.

\paragraph{GPU specifics}

For the training on the datasets {\small\textsc{Mc-Sml-40}},  {\small\textsc{Mc-Big-40}}, {\small\textsc{Ga-10-100}} and {\small\textsc{Ga-20-400}} we use  GPUs Nvidia Quadro RTX 5000 with 16 GB of RAM.
To train the datasets {\small\textsc{Mc-Sml-Var}} and  {\small\textsc{Mc-Big-Var}} we use Nvidia A40 GPUs accelerators with 48Gb of RAM.
To test the performance we use Nvidia A40 GPUs accelerators with 48Gb of RAM for all models and all the datasets on validation and test.

\paragraph{CPU specifics}

The warm starting of the proximal Bundle in SMS++ needs only CPU, the experiments are done on Intel Core i7-8565U CPU @ 1.80GHz × 8. 

\section{k-NN}\label{app:kNN}

We consider the same features as for MLP (see Appendix~\ref{app:mlpfeatures}) and independently select for each dualized constraint $c$ the 20 nearest neighbours with respect to the Euclidean distance. The LM predicted for $c$ is the mean of the LMs associated with its neighbours.
It is important to note that it is a supervised learning method while ours is an unsupervised one.
We tried different values of $k$ from 1 to 20 and we find that the best choice is 20.
For the implementation we use the julia package NearestNeighbors.jl \footnote{\url{https://juliapackages.com/p/nearestneighbors}}.

\section{Features used for MLP and k-NN}
\label{app:mlpfeatures}

Since MLP and k-NN do not use a mechanism such as convolution to propagate the information between the representations of the dualized  constraints, we consider for initial features of each dualized constraint all the information provided to our model (see Appendix~\ref{sec:initial-features} for details), as well as a weighted linear combination of variable feature vectors. The weights are the variable coefficients in that constraint and each feature vector contains the initial features provided to our model for the variable and the following additional information:
\begin{itemize}
    \item the mean values and deviations of the coefficients of that variable on the dualized constraints, and on the non dualized ones,
    \item its lower and upper bounds.
\end{itemize}

\section{Ablation Study - Number of Layers}
\label{app:LayersNumber}

In Tables \ref{tab:layersMC40} and \ref{tab:layersGA}, we present the gaps of three different architectures with varying numbers of layers. The columns represent three different architectures: "Ours," the architecture we introduce in this work; "Nair," the architecture proposed by Nair et al. in \cite{nair2018learning}; and "Gasse," the one presented in \cite{gasse_exact_2019}.
The rows indicate an incremental number of layers from one to ten.

From Table \ref{tab:layersGA}, we observe that for GA, using more than four layers seems to be counterproductive. This can be explained by examining the bipartite graph representation of the instance. In the Generalized Assignment problem, the shortest path between two different nodes associated with the relaxed constraints always consists of four edges. For the Multi-commodity problem, there is no similar bound, as the shortest path between two relaxed nodes depends on the specific structure of the instance. From Table \ref{tab:layersMC40}, we see that adding more than six layers leads to diminishing improvements, though we can still enhance solution quality by increasing the number of layers.


Gasse's architecture is also more unstable, which could result in significantly higher gaps with some layers compared to fewer layers. This instability may be due to the absence of Layer Normalization, leading to very high gradient values.

\begin{table}
\begin{tabular}{r|rrr}
 \# Layers & Ours & Nair & Gasse \\
\hline
 1 & 7.56  & 7.63  &  9.59 \\
 2 & 5.27  & 5.23   & 10.11  \\
 3 & 3.18  & 3.30   & 9.47  \\
 4 & 2.62  & 3.06   & 2.72  \\
 5 & 2.29  & 2.47   & 2.59  \\
 6 & 1.90   &    2.23       &    2.76      \\
 7 & 1.80  &   1.91        &     2.80     \\
 8 & 1.69  & 1.76 &  2.68   \\
 9 & 1.64  &      1.70     &   2.84       \\
 10 & 1.56  &      1.56     &    3.16      \\
\end{tabular}
\caption{ Test set 1 sample - \textsc{Mc-Sml-40} - GAP }
\label{tab:layersMC40}
\end{table}

\begin{table}
\begin{tabular}{r|rrr}
\# Layers & Ours & Nair & Gasse \\
\hline
1 & 0.553 & 0.557 & 0.70 \\
2 & 0.543 & 0.546 & 0.699 \\
3 & 0.533 & 0.524 & 0.695 \\
4 & 0.509 & 0.524 & 0.690 \\
5 & 0.512 &  0.517 & 0.682 \\
6 & 0.513 & 0.518 & 0.720 \\
7 & 0.510 & 0.511 & 0.785 \\
8 & 0.512 & 0.517 & 0.752 \\
9 & 0.511  & 0.515 & 0.785 \\
10 & 0.512 & 0.516 & 0.722 \\
\end{tabular}
\caption{ Test set 1 sample - \textsc{Ga-10-100} - GAP (s) }
\label{tab:layersGA}
\end{table}

{
Notice that the results in Table~\ref{tab:layersMC40} and Table~\ref{tab:layersGA} show small differences compared to the ones on the main part for 5 layers, as they correspond to other runs of the training.
}
\section{Subgradient Method Initialization}
\label{app:SGDinit}

\begin{table*}[t!]
  	\caption{Impact of initialization for a Sub-Gradient solver on \textsc{Mc-Big-Var}.
      We consider initialization from the null vector (zero), the continuous relaxation duals (CR), and our model (Ours).
      We set the maximum iterations to 100000.
}

     \vspace{0.25cm}
 \label{table:SGD}
	\centering
  {\scriptsize
\begin{tabular}{rrrrrrrrr}
 \toprule 
		\multirow{2}{0.5cm}{$\quad \epsilon$} & \multicolumn{2}{c}{zero} & & \multicolumn{2}{c}{CR} &  & \multicolumn{2}{c}{Ours} \\
  \cmidrule{2-3} \cmidrule{5-6}\cmidrule{8-9}
	                                        & time (s)        &    \# iter.   & & time (s)    &   \# iter.      &  &     time (s)     &      \# iter.    \\
		\midrule
1e-1 & 275.09 ($\pm$ 166.59 ) & 86755.53 ( $\pm$ 28222.26 ) & &  284.74 ($\pm$ 184.58) & 85899.10 ($\pm$ 29126.65) & & \textbf{271.73} ($\pm$ 168.97) & \textbf{84882.55} ($\pm$ 29704.22 ) \\
\midrule
1e-2 & 281.00 ($\pm$ 168.63 ) & 88175.07 ( $\pm$ 27438.57 ) & &  291.40 ($\pm$ 186.65) & 87513.27 ($\pm$ 28093.75) & &  \textbf{278.55} ($\pm$ 171.83) & \textbf{86526.59} ($\pm$ 29003.66 )\\
\midrule
1e-3 & 281.00 ($\pm$ 168.64 ) & 88176.12 ( $\pm$ 27439.02 ) & &  291.66 ($\pm$ 186.67) & 87605.24 ($\pm$ 28110.87) & &  \textbf{278.58} ($\pm$ 171.82) & \textbf{86540.76} ($\pm$ 29003.68 ) \\
\midrule
1e-4 &281.00 ($\pm$ 168.64 ) & 88176.12 ( $\pm$ 27439.02 ) & &  291.66 ($\pm$ 186.67) & 87605.24 ($\pm$ 28110.87) & &   \textbf{278.58} ($\pm$ 171.82) & \textbf{86540.76} ($\pm$ 29003.68 )
\\
\bottomrule
\end{tabular}
}
\end{table*}

In Table~\ref{table:SGD} we perform the same experiments as Table~\ref{table:Bundle} with a solver implementing the subgradient method.
We see that 
{
subgradient method requires much more time than the bundle method, even for low precision levels.
The initialization yields more or less the same results. This is likely due to the step size scheduler, which always starts with a step size of one. Then, at a given iteration $i$, the learning rate is $\frac{1}{1+m}$, where $m$ is the total number of iterations where the predicted value is worse than the previous iteration.
An accurate choice of step size can lead to better results, particularly for non-zero initialization. However, this should be done specifically for each initialization (and possibly for each instance), which is beyond the scope of this work.
}

\end{document}